\documentclass[10pt,twocolumn,letterpaper]{article}

\usepackage[pagenumbers]{iccv} 
\usepackage{algorithm}
\usepackage{algpseudocode}
\usepackage{amsmath}
\usepackage{booktabs}
\usepackage{multirow}
\usepackage{array}
\usepackage{makecell}
\usepackage{mathrsfs}
\usepackage{chngcntr} 
\usepackage{ragged2e}

\definecolor{iccvblue}{rgb}{0.21,0.49,0.74}
\usepackage[pagebackref,breaklinks,colorlinks,allcolors=iccvblue]{hyperref}

\def\paperID{6280} 
\def\confName{ICCV}
\def\confYear{2025}

\title{Dataset Distillation as Data Compression: A Rate-Utility Perspective
}

\author{Youneng Bao$^{1,3,}$\thanks{Equal contribution.}, \ Yiping Liu$^{1,}$\footnotemark[1], \ Zhuo Chen$^{2}$, \  Yongsheng Liang$^{1}$, \ Mu Li$^{1,}$\thanks{Corresponding authors.}, \ Kede Ma$^{3,}$\footnotemark[2] \\
$^{1}$Harbin Institute of Technology, Shenzhen\ \ \ $^{2}$Peng Cheng Laboratory 
\ \ $^{3}$City University of Hong Kong
\\
{\tt\small 
younebao@cityu.edu.hk, yipingliu@stu.hit.edu.cn, chenzh08@pcl.ac.cn} \\
\tt\small  {\{liangys,limu2022\}@hit.edu.cn,
kede.ma@cityu.edu.hk} \\
\tt\small \url{https://nouise.github.io/DD-RUO}
}

\begin{document}
\maketitle

\begin{abstract} 

 Driven by the ``scale-is-everything'' paradigm, modern machine learning increasingly demands ever-larger datasets and models, yielding prohibitive computational and storage requirements. Dataset distillation mitigates this by compressing an original dataset into a small set of synthetic samples, while preserving its full utility. Yet, existing methods either maximize performance under fixed storage budgets or pursue suitable synthetic data representations for redundancy removal, without jointly optimizing both objectives. In this work, we propose a joint rate-utility optimization method for dataset distillation. We parameterize synthetic samples as optimizable latent codes decoded by extremely lightweight networks. We estimate the Shannon entropy of quantized latents as the rate measure and plug any existing distillation loss as the utility measure, trading them off via a Lagrange multiplier. To enable fair, cross-method comparisons, we introduce bits per class (bpc), a precise storage metric that accounts for sample, label, and decoder parameter costs. On CIFAR-10, CIFAR-100, and ImageNet-128, our method achieves up to $170\times$ greater compression than standard distillation at comparable accuracy. Across diverse bpc budgets, distillation losses, and backbone architectures, our approach consistently establishes better rate-utility trade-offs.

\end{abstract}

\section{Introduction}
\label{sec:intro}

 The rapid advances in machine learning have been driven by the ``scale-is-everything'' paradigm~\cite{Scaling2017,Kaplan2020ScalingLF}, in which ever-larger models trained on ever-more data yield progressively better performance across various computational prediction tasks~\cite{brown2020language,DosovitskiyB0WZ21}. However, this scaling trend incurs steep computational, storage, and environmental costs, posing a barrier to sustainable and accessible research and deployment. Dataset distillation (DD)~\cite{wang2018dataset} addresses these challenges by learning a small, synthetic dataset that enables rapid model prototyping, accelerated model training, and efficient hyperparameter tuning, while lowering compute, storage, and energy requirements~\cite{lei2023comprehensive,liu2025evolution}. 

\begin{figure}[t]
    \centering
    \includegraphics[width=0.95\linewidth]{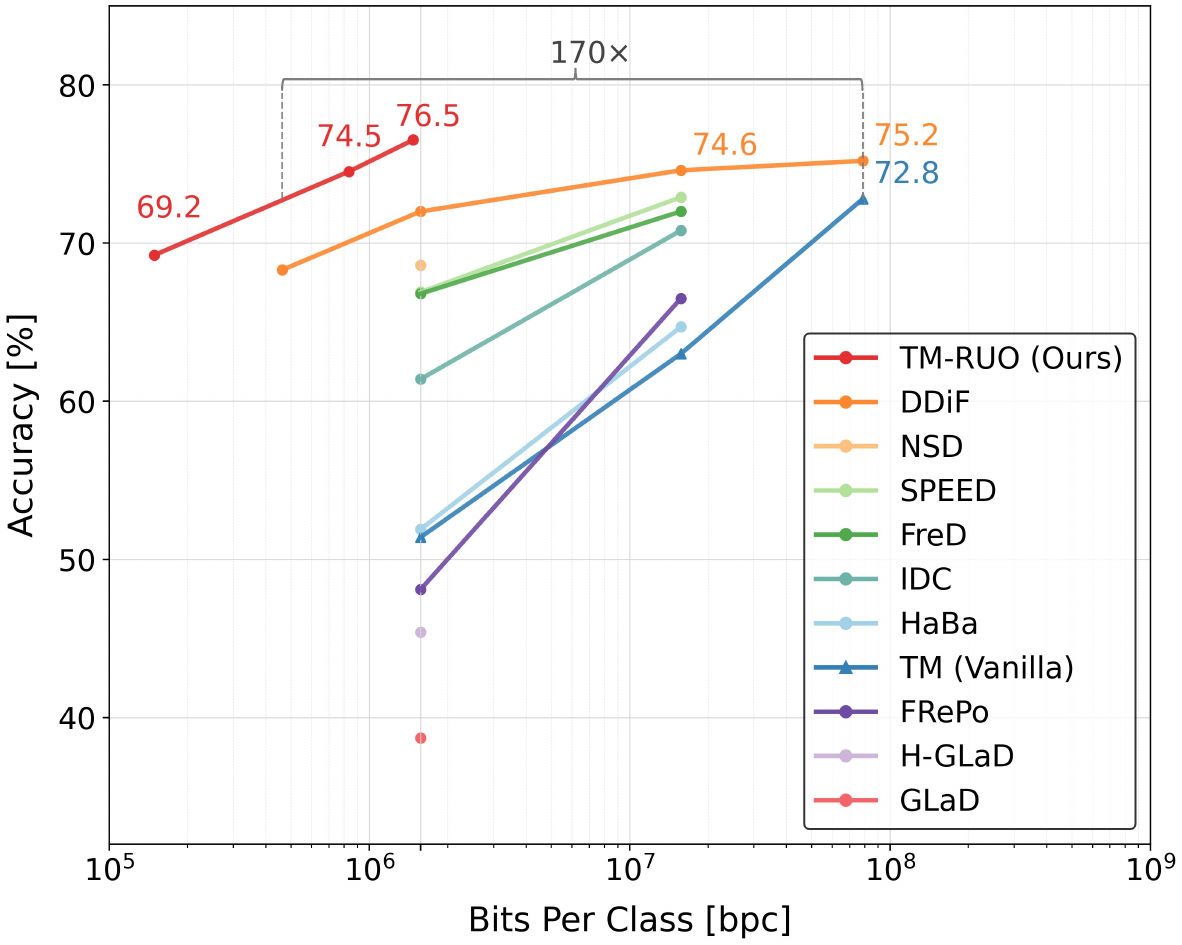}
    \caption{Comparison of the rate-utility curves on the Nette subset of ImageNet, with the rate axis displayed on a logarithmic scale. The proposed TM-RUO that integrates the trajectory matching loss~\cite{cazenavette2022dataset} consistently achieves superior accuracy across a wide range of storage budgets, as measured in bits per class (bpc).}
    \label{fig:ru_nette}
\end{figure}

Initially, DD is formulated as a bilevel optimization problem~\cite{wang2018dataset}, necessitating computationally intensive, memory-hungry backpropagation through time (BPTT), which is also prone to gradient instability~\cite{58337}. To alleviate this, meta-gradient computation has been pursued or bypassed through a variety of techniques, including truncated BPTT~\cite{feng2024embarrassingly}, closed-form kernel ridge regression~\cite{nguyen2021datasetmeta}, gradient and distribution matching~\cite{zhao2020dataset,Zhao_2023_WACV}, and 
implicit differentiation~\cite{LooHLR23}. 
Meanwhile, representation-based methods~\cite{pmlr-v162-kim22c,shin2023frequency,wei2023sparse,Cazenavette2023GeneralizingDD,yang2024neural,zhang2025td3} mitigate redundancy in synthetic data by parameterizing it as low-dimensional latent codes, which can be reconstructed into full-sized images via structured or generative decoders.

At the core of DD lies a trade-off between the \textit{rate} (\ie, the storage footprint of synthetic datasets) and \textit{utility} (\ie, the performance of models trained on distilled data). 
Despite impressive progress, existing DD methods optimize rate and utility separately, precluding Pareto-optimal trade-offs. Inspired by the rate-distortion theory that governs data compression~\cite{book2006tit,Ball2016EndtoendOI}, we propose a joint rate-utility optimization method for DD. We parameterize synthetic samples as a set of multiscale optimizable latent codes that can be decoded by extremely lightweight networks. To compute the distillation \textit{rate}, we model each \textit{quantized} latent code with a context-aware Laplace distribution whose mean and scale parameters are inferred from its causal neighbors~\cite{kim2024c3}. We then approximate the rate using the Shannon entropy under this Laplace prior. To assess the distillation \textit{utility}, we leverage the plug-and-play property of our method by incorporating any off-the-shelf distillation loss~\cite{zhao2020dataset,cazenavette2022dataset,Zhao_2023_WACV}. Finally, we optimize all parameters end-to-end against the joint rate-utility objective, balancing the two terms via a tunable Lagrange multiplier~\cite{Ball2016EndtoendOI}.

To facilitate equitable comparisons across DD methods of different design philosophies, we introduce the bits per class (bpc) metric, which quantifies the average number of bits needed to store a distilled dataset---including sample representations, class labels, and decoder parameters (if any)---on a per-class basis (see Fig.~\ref{fig:ru_nette}).  In contrast, the widely adopted images per class (ipc) metric scales linearly with raw image dimensions and overlooks both the bit cost of encoding labels and decoder parameters.

In summary, our key contributions include 
\begin{itemize}
\item A joint rate-utility optimization method for DD that retains full compatibility with existing distillation losses;
\item A bpc metric that enables standardized and fair rate-utility comparisons across various DD methods;
\item An experimental demonstration on CIFAR-10, CIFAR-100, and ImageNet-128  that our method achieves better rate-utility trade-offs across various storage budgets, distillation losses, and network architectures.
\end{itemize}

\section{Related Work}
\label{sec:related}
In this section, we review prior efforts in DD and data compression. Our method bridges the conceptual and computational gap between these two fields, allowing principled exploration of Pareto-optimal trade-offs between the rate and utility in DD. 

\subsection{Dataset Distillation (DD)}
\label{sec:sub_dd}
Wang~\etal~\cite{wang2018dataset} first formulated DD as a bilevel optimization problem, which requires BPTT, incurring prohibitive computation due to long unrolls and nested differentiation. 
To reduce this burden, several meta-gradient estimation strategies have been proposed. Feng~\etal~\cite{feng2024embarrassingly} introduced random truncated BPTT to cap the number of unrolled steps, while Nguyen~\etal~\cite{nguyen2021dataset} replaced the iterative inner-loop solver with closed-form kernel ridge regression. 

Intuitively, aligning the training dynamics of original and synthetic datasets provides more granular supervisory signals than merely measuring final performance. As specific instances, gradient matching~\cite{zhao2020dataset} aligns instantaneous parameter updates, whereas trajectory matching~\cite{cazenavette2022dataset} extends this alignment across the entire sequence. By aligning distributions in a predefined feature space, distribution matching \cite{Zhao_2023_WACV} bypasses the challenge of measuring functional similarity in parameter space and avoids costly second-order gradient computations.

More recent work focuses on decoupling or reframing the bilevel structure to handle large-scale, high-resolution datasets. The Teddy method~\cite{YuLYW24} applies first-order Taylor expansion to convert nested gradients into sums of inner products, equivalent to matching feature-space means and variances under mild assumptions. SRe2L~\cite{yin2023squeeze} decomposes DD into three sequential stages: 1) train a teacher network on original data, 2) optimize synthetic samples by matching the BatchNorm statistics and outputs against the frozen teacher, and 3) relabel the synthetic samples.

In parallel, representation-based DD methods have been devised to reduce the storage footprint of synthetic data. Explicit (\ie, code-centric) approaches learn low-dimensional (latent) codes with fixed decoders---implemented via discrete cosine transform (DCT)~\cite{shin2023frequency}, Tucker decomposition~\cite{zhang2025td3}, or deep generative priors such as generative adversarial networks (GANs)~\cite{Cazenavette2023GeneralizingDD} and diffusion models~\cite{zhong2024hierarchical,Su2024D4MDD,chen2025influenceguided}\footnote{The vanilla DD~\cite{wang2018dataset} can also be regarded as an explicit scheme, where the decoder reduces to the identity mapping.}. Implicit (\ie, decoder-centric) methods encode synthetic samples solely in decoder parameters given random latent inputs~\cite{shin2025distilling}. Hybrid schemes optimize both latents and decoders for improved performance~\cite{deng2022remember,liu2022dataset}. 

However, to the best of our knowledge, no existing DD method formalizes storage cost using information-theoretic rate measures (\eg, Shannon entropy), nor incorporates such rate considerations into a joint end-to-end optimization of storage and utility in DD.

\subsection{Data Compression}
Data compression is generally divided into two complementary paradigms: lossless and lossy compression. Lossless methods aim to encode data exactly, exploiting statistical redundancies via entropy coding (\eg, Huffman coding~\cite{huffman1952method} and arithmetic coding~\cite{witten1987arithmetic}) in concert with probability models. Classic schemes such as dictionary-based algorithms (\eg, LZW~\cite{welch1984technique}) and transform coding remain popular in applications for their reliability and low complexity. Modern approaches increasingly integrate learnable generative models to capture high-dimensional, long-range dependencies: Auto-regressive neural networks~\cite{UriaML13}, latent-variable models~\cite{kingma2013auto}, and invertible flow architectures~\cite{DinhKB14} can be paired with entropy coders to approach the information-theoretic limits~\cite{yang2023introduction}.

By contrast, lossy compression introduces a controlled distortion to achieve greater bitrate savings. Rate-distortion theory formalizes this trade-off by minimizing a weighted sum of expected bitrate and distortion~\cite{shannon1959coding}. Traditional pipelines apply fixed linear transforms (such as DCT or wavelets~\cite{dct1974}), followed by quantization and entropy coding. Neural lossy compression~\cite{Ball2016EndtoendOI} replaces these hand-crafted transforms with trainable encoders and decoders\footnote{In signal processing, the encoder and decoder are referred to as analysis and synthesis transforms, respectively.}, augmented by learned quantization schemes and by rich entropy models combining hyperpriors~\cite{ballé2018variational} with autoregressive~\cite{MinnenBT18} and discretized mixture-of-likelihood~\cite{li2020efficient} contexts. Moreover, advanced transform modules (\eg, generalized divisive normalization~\cite{BalleLS15}, residual connection~\cite{he2016deep}, and non-local attention~\cite{VaswaniSPUJGKP17}) for improved latent representation,
GAN-based distortion measures for perceptual realism~\cite{mentzer2020high}, and diffusion-based decoders for fine-grained reconstruction~\cite{agustsson2023multi,yang2023lossy}  have each pushed the frontier of rate-distortion performance.

Instead of relying on a single, shared decoder, recent ``overfitted'' compression methods~\cite{Dupont2021COINCW,guo2023compression,kim2024c3} tailor both the synthesis and entropy networks to each image, storing their quantized weights alongside the latent codes. 
Although it incurs per-sample training overhead, this approach produces neural codecs that are both highly efficient and straightforward to deploy, striking a compelling balance between compression efficacy and run-time simplicity. Consequently, it is especially well suited for representing synthetic data and for obtaining precise rate measurements in DD, as demonstrated in our study.

Although DD and data compression aim for distinct goals, they both seek
 minimum-entropy representations that preserve only the information essential for downstream tasks. We exploit this commonality by embedding differentiable quantization, entropy modeling, and rate-distortion optimization from neural lossy compression into our method, enabling joint end-to-end optimization of rate and utility in DD.

\section{Proposed Method}
\label{sec:Methodology}
\begin{figure*}[t]
    \centering    \includegraphics[width=0.99\linewidth]{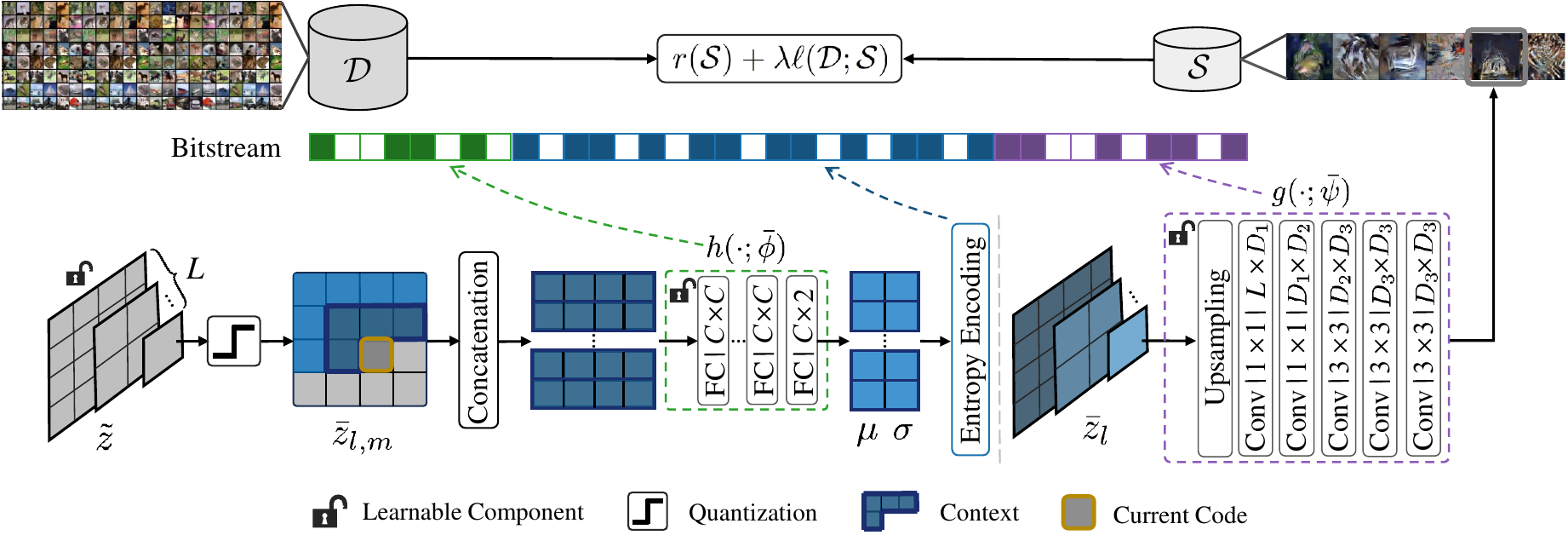}
    \caption{System diagram of our proposed joint rate-utility optimization method for DD. ``FC'' denotes a fully connected layer whose parameter dimensions are given by the product of input width and output width, and $C$ represents the context length. Convolution layers are specified using the format: ``kernel height $\times$ kernel width $\vert$ input channels $\times$ output channels.''
    }
    \label{fig:our_framework}
\end{figure*}

In this section, we present our DD method through the lens of joint rate-utility optimization (see Fig.~\ref{fig:our_framework}). We first introduce preliminaries in Sec.~\ref{subsec:pre} and give our general problem formulation in Sec.~\ref{subsec:profor}. We then derive a differentiable rate term tailored to synthetic datasets and embed it in our joint objective in Secs.~\ref{subsec:dsd} and~\ref{subsec:jruo}. Finally, Sec.~\ref{subsec:bpc} defines a precise bpc metric for standardized evaluation.

\subsection{Preliminaries}\label{subsec:pre}
Let $\mathcal{D} =\{ (x^{(i)}, y^{(i)}) \}_{i=1}^{M}$ denote the original dataset of $M$ labeled examples and let  $\mathcal{S} =\{ (\bar{x}^{(i)}, \bar{y}^{(i)}) \}_{i=1}^{N}$ be a much smaller synthetic dataset with $N \le M$. The goal of DD is to learn $\mathcal{S}$ so that a model trained on $\mathcal{S}$ achieves performance close to one trained on $\mathcal{D}$. Traditionally, this is cast as a bilevel optimization problem:
\begin{equation}\label{eq:ddbf}
\begin{aligned}
&\min_{\mathcal{S}}\;\ell_\mathrm{outer}\left(\mathcal{D};\theta^\star(\mathcal{S})\right)\\
&\text{subject to}\; 
\theta^\star(\mathcal{S}) = \arg\min_{\theta}\;\ell_\mathrm{inner}(\mathcal{S};\theta),
\end{aligned}
\end{equation}
where 
\begin{equation}
\ell_\mathrm{inner}(\mathcal{S};\theta)=\mathbb{E}_{(\bar x, \bar y)\sim \mathcal{S}}\left[
\ell_\mathrm{inner}\left(f\left(\bar x;\theta\right),\bar y\right)\right]
\end{equation}
is the training loss (\eg, cross-entropy) on the synthetic dataset, and
$f(\cdot;\theta)$ is a differentiable classifier (\eg, a neural network), parameterized by $\theta$.  $\theta^\star(\mathcal{S})$ is typically found by unrolling $T$ gradient steps, with an initial guess $\theta^{(0)}$:
\begin{align}
   \theta^{(t)} = \theta^{(t-1)}  - \eta\nabla_\theta\ell_\mathrm{inner}(\mathcal{S};\theta^{(t-1)}),\, t = 1,\dots,T,
\end{align}
where $\eta$ is the learning rate and $\theta^{(T)}$ serves as an approximation of the true optimum $\theta^\star(\mathcal{S})$. Finally, $\ell_\mathrm{outer}$ measures how well $f(\cdot;\theta^\star(\mathcal{S}))$ generalizes to the original data $\mathcal{D}$.

Solving Problem~\eqref{eq:ddbf} via BPTT incurs substantial computational and memory overhead. To alleviate this, one can truncate the iterative inner-loop solver~\cite{feng2024embarrassingly}, replace it with closed-form surrogates~\cite{nguyen2021dataset}, or decouple the dependence of the optimal parameters $\theta^\star$ from the synthetic dataset $\mathcal{S}$~\cite{YuLYW24,yin2023squeeze}. To reduce storage requirements, the synthetic dataset itself may be parameterized by a collection of latent codes $\mathcal{Z} = \{\bar z^{(i)}\}_{i=1}^{N}$ and class labels $\mathcal{Y} = \{\bar y^{(i)}\}_{i=1}^{N}$, 
together with lightweight decoders $g(\cdot, \Psi)$ with parameters $\Psi = \{
\bar{\psi}^{(i)}\}_{i=1}^{N}$:
\begin{equation}\label{eq:hp}
    \mathcal{S}\left(\mathcal{Z}, \mathcal{Y}, \Psi\right) = \left\{\left(g\left(\bar z^{(i)};\bar{\psi}^{(i)}\right), \bar y^{(i)}\right)  \right\}_{i = 1}^{N}.
\end{equation}
Prior methods enforce a hard constraint on pixel-based storage budget (\ie, $\# \text{pixels} \times \text{bit-depth}$) but do not explicitly model the information-theoretic cost of encoding these latents, labels, and decoder parameters.

\subsection{Problem Formulation}\label{subsec:profor}
Inspired by the rate-distortion theory~\cite{shannon1959coding,book2006tit}, we propose to cast DD as a joint rate-utility optimization problem. Specifically, we introduce a differentiable rate term $r(\mathcal{S})$,
and our objective balances this cost against any choice of distillation loss $\ell(\mathcal{D};\mathcal{S})$ on the original data:
\begin{equation}\label{eq:probform}
    \min_\mathcal{S}\; r(\mathcal{S}) + \lambda \ell(\mathcal{D};\mathcal{S}),
\end{equation}
where $\lambda > 0$ governs the trade-off between the two terms. In this study, we instantiate $\ell(\mathcal{D};\mathcal{S})$ in three different ways.
\begin{itemize}
    \item \textit{Gradient matching}~\cite{zhao2020dataset}. We align minibatch gradients computed on $\mathcal{S}$ with those on $\mathcal{D}$ at each training step:
    \begin{equation}
\ell_\mathrm{grad}(\mathcal{D};\mathcal{S})=\sum_{t=0}^T\left\Vert\nabla_\theta\ell\left(\mathcal{D};\theta^{(t)}\right) -\nabla_\theta\ell\left(\mathcal{S};\theta^{(t)}\right)\right\Vert_2^2,
\end{equation}
where $\{\theta^{(t)}\}_{t=0}^T$ are obtained by training on $\mathcal{S}$.
\item \textit{Trajectory matching}~\cite{cazenavette2022dataset}. We encourage the training trajectory on $\mathcal{S}$ to follow an ``expert'' trajectory on $\mathcal{D}$:
\begin{equation}
\ell_\mathrm{traj}(\mathcal{D};\mathcal{S})=\sum_{t=0}^T\frac{\left\Vert  \theta^{(t + t_2)}(\mathcal{D}) - \theta^{(t+t_1)}(\mathcal{S})\right\Vert_2^2}{\left\Vert  \theta^{(t+t_2)}(\mathcal{D}) - \theta^{(t)}(\mathcal{D})\right\Vert_2^2},
\end{equation}
where $t_1 < t_2$ and the denominator, as a form of normalization, ensures that the loss measures relative alignment rather than absolute magnitude.
\item \textit{Distribution matching}~\cite{Zhao_2023_WACV}. We match the feature-space distributions over $\mathcal{D}$ and $\mathcal{S}$:
\begin{align}
\ell_\mathrm{dist}(\mathcal{D};\mathcal{S})= 
d\left(
  \mathbb{E}_{x\sim\mathcal{D}}\left[
    \varphi({x})\right],
  \;
  \mathbb{E}_{\bar x\sim\mathcal{S}}\left[
    \varphi(\bar x)\right]
\right),
\end{align}
where $\varphi(\cdot)$ denotes a fixed feature extractor and $d(\cdot, \cdot)$ is a statistical distance (\eg, maximum mean discrepancy~\cite{GrettonBRSS12}).
\end{itemize}

By integrating rate and utility into a single end-to-end objective, our method enables systematic exploration of distilled datasets that optimally trade off storage cost against generalization performance.

\subsection{Rate Modeling}\label{subsec:dsd}
In our joint rate-utility optimization method, we employ a hybrid parameterization~\cite{liu2022dataset,ladune2023cool} of the synthetic dataset $\mathcal{S}(\mathcal{Z},\mathcal{Y},\Psi)$ in Eq.~\eqref{eq:hp}, where each synthetic sample $\bar x^{(i)}\in\mathcal{S}$ is generated from a set of latent codes
 at $L$ scales $\{\bar z^{(i)}_l\}_{l=1}^{L}\subset\mathcal{Z}$ via a lightweight decoder $g\left(\cdot;\bar{\psi}^{(i)}\right)$ with $\bar{\psi}^{(i)}\subset\Psi$ and annotated with a class label $\bar y^{(i)}\in\mathcal{Y}$. We  decompose the total bitrate into three additive terms:
\begin{align}
  r(\mathcal{S}) =   r(\mathcal{Z}) + r(\mathcal{Y}) + r(\Psi).
\end{align}
Below, we describe how each term is modeled.


\noindent\textbf{Bitrate for Latent Codes.} 
We arrange the \textit{continuous} latent code $\tilde z$ at $L$ scales\footnote{To improve notational clarity, we omit the superscript image index from the mathematical expressions.} so that $\tilde z = \{\tilde z_l\}_{l=1}^{L}$, where $\tilde z_l\in\mathbb{R}^{\left\lfloor\frac{H}{2^{l-1}}\right\rfloor\times \left\lfloor\frac{W}{2^{l-1}}\right\rfloor}$, and $H$ and $W$ are the height and width of the original image. To enable entropy coding,  the $m$-th latent code at the $l$-th scale, $\tilde{z}_{l,m}$, is first quantized by rounding to the nearest integer:
\begin{equation}
    \bar{z}_{l,m} = \mathrm{round}(\tilde{z}_{l,m}) = \left\lfloor\tilde z_{l,m} + \frac{1}{2}\right\rfloor.
\end{equation}
We then fit a lightweight, sample-specific, and spatially autoregressive entropy model:
\begin{equation}
    P(\bar{z}_{l,m}\vert \bar{c}_{l,m}) = \int_{\bar z_{l,m}-\frac{1}{2}}^{\bar z_{l,m}+\frac{1}{2}} p(\xi\vert\bar{c}_{l,m})\,\mathrm{d}\xi,
\end{equation}
where the context $\bar{c}_{l,m}$ aggregates previously decoded neighbors, and
\begin{equation}
    p(\xi\vert \bar{c}_{l,m})=\mathrm{Laplace}(\xi;\mu_{l,m},\sigma_{l,m}),
\end{equation}
with $\mu_{l,m}, \sigma_{l,m} = h(\bar{c}_{l,m};\phi)$ 
produced by an auxiliary entropy network with \textit{continuous} parameters $\phi$. Under the assumption of independence across channels and scales, the bitrate of the latent codes $\mathcal{Z}$ can be computed by 
\begin{align}
\label{eq:rate of latent}
r\left(\mathcal{Z};\Phi\right)
    = -\frac{1}{N}\sum_{i=1}^N\sum_{l,m} \log_2 P\left(\bar z^{(i)}_{l,m}\middle\vert \bar{c}^{(i)}_{l,m}\right).
\end{align}

To account for the additional storage overhead of the entropy network parameters $\phi$, which are represented as $32$-bit floats during training, we apply a uniform, post-training quantization, as suggested in~\cite{ladune2023cool,kim2024c3}. A step size $Q_e$ is selected by grid-search to balance bitrate against reconstruction error (as measured by the mean squared error). Concretely, the $j$-th parameter $\phi_j$ is quantized as
\begin{align}\label{eq:round_phi}
\bar{\phi}_j = q\left(\phi_j;Q_e\right)=\mathrm{round}\left(\frac{\phi_j}{Q_e}\right) \times Q_e.
\end{align}
The quantized weights $\bar{\phi}\subset\Phi$ are then encoded losslessly under a discretized, zero-mean Laplace prior, whose scale is set to $\mathrm{std}(\bar{\phi})/\sqrt{2}$. Averaging over $N$ entropy network instances, the resulting  bitrate becomes
\begin{equation}
\label{eq:rate of entropy net}
    r(\Phi) = -\frac{1}{N}\sum_{i=1}^N \sum_j \log_2 P\left(
    \bar{\phi}^{(i)}_j\right).
\end{equation}

\begin{algorithm}[t]
\caption{Joint Rate-Utility Optimization for DD}
\label{alg:Fred_revised}
\begin{algorithmic}[1]
    \Statex \colorbox{gray!30}{\parbox{\dimexpr\linewidth-2\fboxsep}{{\textit{Phase 1: Initialization}}}}
    \State Randomly initialize $\{\tilde{z}^{(i)}, \phi^{(i)}, \psi^{(i)}, \bar{y}^{(i)}\}_{i=1}^{N}$ 
    \For{$i = 1$ to $N$} \Comment{in parallel}
        \State Randomly sample an original pair $({x},{y})$ in $\mathcal{D}$ 
            \State Optimize $\{\tilde{z}^{(i)},\phi^{(i)},\psi^{(i)}\}$ by solving \cref{eq:init}
             \State $\bar{y}^{(i)} \leftarrow y$
        \EndFor
    \Statex \colorbox{gray!30}{\parbox{\dimexpr\linewidth-2\fboxsep}{\textit{Phase 2: Joint Rate-Utility Optimization}}}    
    \Repeat
        \State \begin{tabular}[t]{@{}l@{}}
            Sample an original mini-batch from $\mathcal{D}$ and a synth-\\
            etic mini-batch from $\mathcal{S}(\mathcal{Z}, \Phi, \Psi, \mathcal{Y})$
        \end{tabular}
        \State \begin{tabular}[t]{@{}l@{}}
        Compute the total rate-utility loss on the sampled\\
        mini-batches using Eq. \eqref{eq:simob}
        \end{tabular}
        \State Update parameters $\{\mathcal{Z}, \Phi, \Psi\}$ via gradient descent
    \Until{convergence}  
    \Statex \colorbox{gray!30}{\parbox{\dimexpr\linewidth-2\fboxsep}{\textit{Phase 3: Post-Quantization}}}
    \State Apply post-quantization to the parameters of the entropy and decoder networks
\end{algorithmic}
\end{algorithm}

\noindent\textbf{Bitrate for Class Labels.} 
Let $\mathcal{Y} = \{\bar{y}^{(i)}\}_{i=1}^N$ denote the class labels of our $N$ synthetic samples, each taking one of $K$ discrete categories with probabilities $\{\bar{y}^{(i)}_k\}_{k=1}^K$, where $\bar{y}^{(i)}_k \ge 0$ and $\sum_k \bar{y}^{(i)}_k = 1$. Under Shannon's source coding theorem~\cite{shannon1948coding}, the minimal bitrate required to losslessly encode $\mathcal{Y}$ is upper-bounded by
\begin{equation}
    r(\mathcal{Y}) < \mathrm{Entropy}(\bar{y}) + 1 \le \log_2 K + 1,
\end{equation}
where $\mathrm{Entropy}(\bar{y}) = -\sum_k \bar{y}_k \log_2 \bar{y}_k$ is the Shannon entropy, and equality in the upper bound holds when the labels are uniformly distributed across all $K$ classes. In many DD methods~\cite{CuiWSH23, yin2023squeeze}, however, one deliberately employs \textit{soft} labels,  that is, probability vectors $\tilde{y} = [\tilde{y}_1, \ldots, \tilde{y}_K]$ lying in the simplex $\Delta^{K-1}$. Because these vectors are continuous, vector quantization~\cite{1162229} should first be applied before entropy coding. For example, to guarantee that each coordinate $\tilde{y}_k$ is approximated to within $\epsilon$, the $(K-1)$-simplex can be uniformly partitioned into on the order of $\epsilon^{-(K-1)}/(K-1)!$ bins. Hence, the resulting  bitrate obeys
\begin{equation}
\label{eq:rate of y}
    r(\mathcal{Y}) \lesssim \log_2 (\# \text{bins})  = -\log_2 (K-1)! - (K-1)\log_2 \epsilon,
\end{equation}
which makes it clear that storing soft labels at full floating-point precision can incur a dramatically higher cost than simply recording hard labels---and in extreme cases may even exceed the cost of transmitting the latent codes~\cite{liu2025evolution} (see Appendix~\ref{subsec:bsl} for detailed derivation).

\noindent\textbf{Bitrate for Decoder Parameters.}
The decoder $g(\cdot;\psi)$ reconstructs each synthetic image $\bar x$ from its multiscale latent codes. Concretely, the latents $\{\bar z_l\}$ are first upsampled to the original resolution and concatenated into a tensor $\mathrm{Up}(\bar z)\in\mathbb{R}^{H\times W\times L}$. $g(\cdot;\psi)$ then maps $\mathrm{Up}(\bar z)$ to $\bar x$, which is subsequently learned by the downstream classifier $f(\cdot;\theta)$ for utility measurement.

\begin{figure*}[t]
  \centering
  \begin{subfigure}[b]{0.495\linewidth}
    \includegraphics[width=\linewidth]{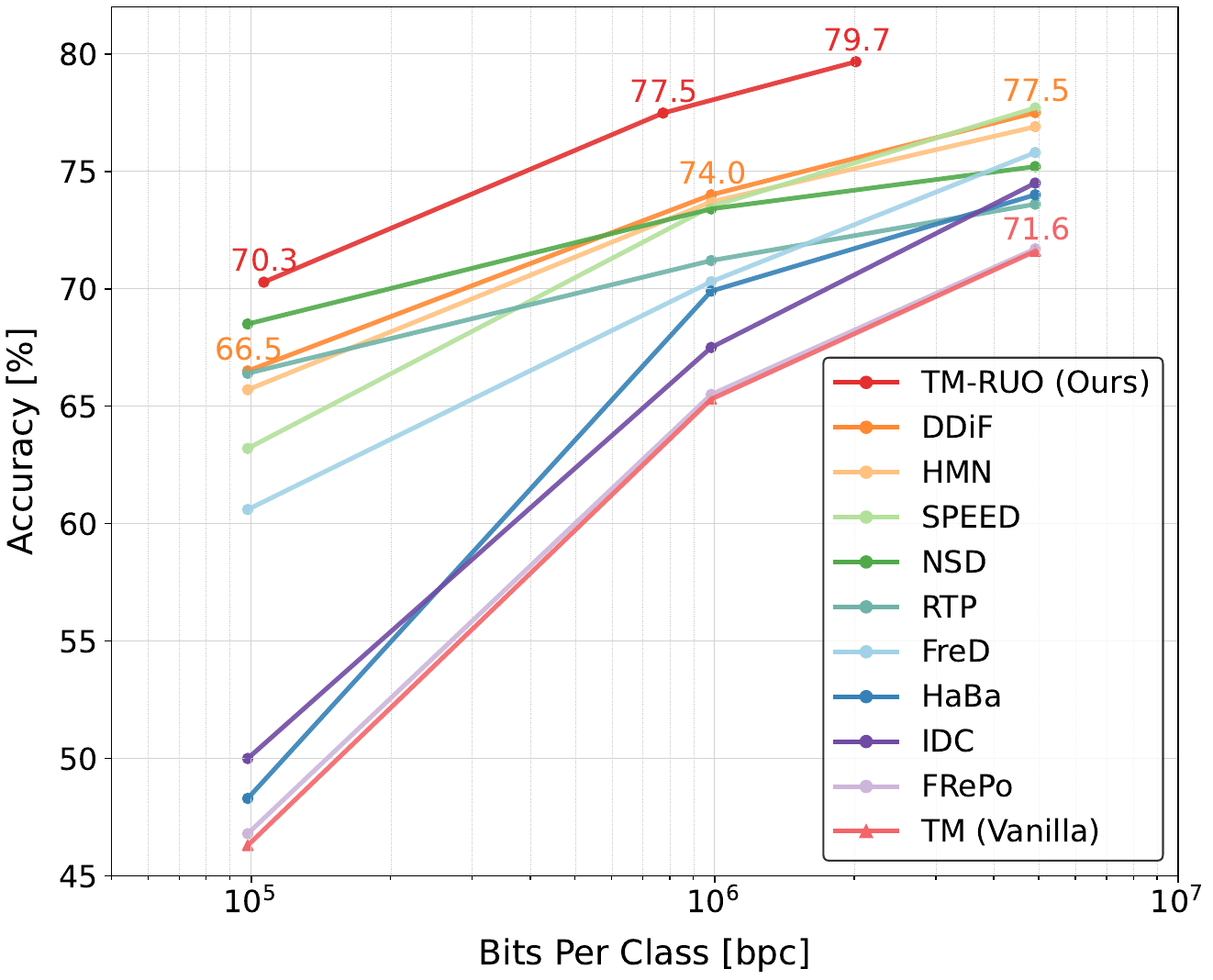}
    \caption{CIFAR-10}
  \end{subfigure}
  \begin{subfigure}[b]{0.495\linewidth}    \includegraphics[width=\linewidth]{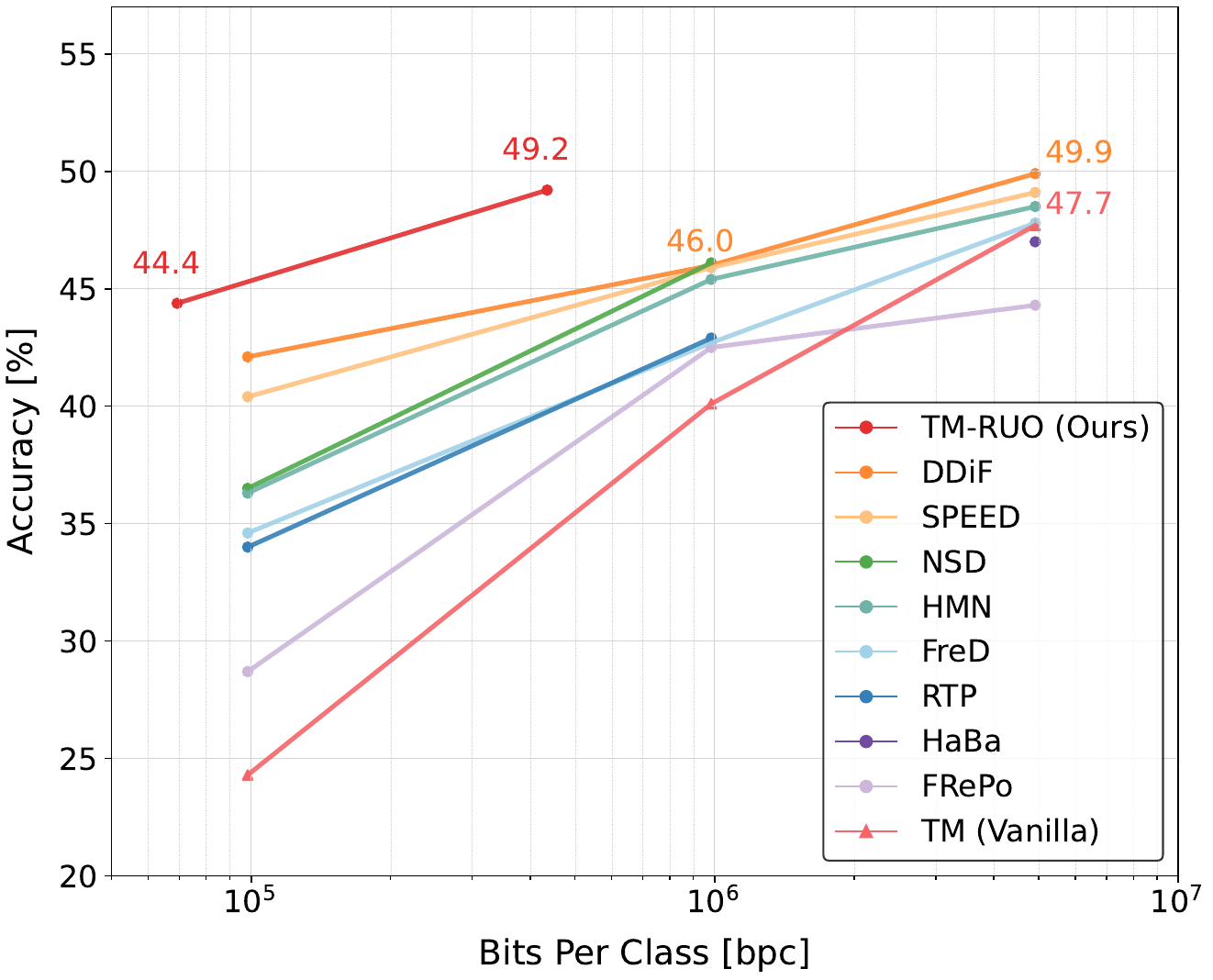}
    \caption{CIFAR-100}
  \end{subfigure}
  \caption{Comparison of the rate–utility curves on CIFAR-10 and CIFAR-100. Our proposed joint rate–utility optimization method achieves superior accuracy across a wide range of bpc budgets.}
  \label{fig:RU}
\end{figure*}

To render the decoder parameters $\psi$ amenable to entropy coding, we apply the same uniform quantization used for the entropy network parameters, \ie, $\bar{\psi}_j = q(\psi_j; Q_d)$, where the quantization step size $Q_d$ is chosen by grid search. Averaging over $N$ decoder instances under a discretized, zero-mean Laplace prior, the resulting  bitrate is 
\begin{equation}
\label{eq:rate of decoder network}
    r(\Psi) = - \frac{1}{N}\sum_{i=1}^N\sum_{j}\log_2 P\left(
    \bar{\psi}^{(i)}_j\right).
\end{equation}
\subsection{Joint Rate-Utility Optimization}\label{subsec:jruo}
Plugging the decomposed rate terms in Eqs.~\eqref{eq:rate of latent},~\eqref{eq:rate of entropy net},~\eqref{eq:rate of y}, and~\eqref{eq:rate of decoder network} into the general formulation in Eq.~\eqref{eq:probform}, we can write the full joint rate-utility objective as
\begin{equation}
\min_{\mathcal{S} = \{\mathcal{Z},\mathcal{Y},\Phi,\Psi\}} r(\mathcal{Z};\Phi) + r(\mathcal{Y}) + r(\Phi) + r(\Psi) + \lambda\ell(\mathcal{D};\mathcal{S}).
\end{equation}
To overcome the non-differentiability of quantization and enable end-to-end optimization, several continuous relaxations can be leveraged, including the straight-through estimator~\cite{BengioLC13}, additive uniform noise injection~\cite{Ball2016EndtoendOI}, and soft (annealed) quantization~\cite{AgustssonMTCTBG17}. In our current implementation, we specialize this to a simplified objective:
\begin{equation}\label{eq:simob}
\min_{\mathcal{Z},\Phi,\Psi}  r(\mathcal{Z};\Phi)+ \lambda\ell(\mathcal{D};\mathcal{Z},\Phi,\Psi),
\end{equation}
omitting $r(\mathcal{Y})$, $r(\Phi)$, and $r(\Psi)$ from joint optimization for the following reasons. First, we employ hard (one-hot) labels for $\mathcal{Y}$, which can be treated as a constant additive offset. Second, both the entropy and decoder networks are architected to be extremely lightweight. We quantize and entropy-encode their weights offline---via a small grid search on uniform step sizes---and thereby fix $r(\Phi)$ and $r(\Psi)$ as constant rate offsets. This decoupling avoids backpropagating through network quantization and considerably reduces computational overhead and gradient variance. Algorithm~\ref{alg:Fred_revised} sketches the entire procedure.

\subsection{Bits Per Class (bpc) Metric}\label{subsec:bpc}
\vspace{-0.15em}
The widely used ipc metric---counting the number of distilled images per category---fails to capture the true storage footprint of a synthetic dataset. First, raw-pixel storage grows linearly with image height, width, channel count, and bit-depth; hence, a fixed ipc can correspond to vastly different bitrates across datasets or methods. Second, ipc ignores any auxiliary overhead, such as the bits required to encode decoder parameters or soft-label representations. 

To provide a unit-consistent, information-theoretic measure of storage cost, we define the bpc metric. Let $\# \text{bits}(\mathcal{S})$ be the total number of bits needed to (losslessly) encode the distilled dataset $\mathcal{S}$. If there are $K$ distinct classes, then 
\begin{equation}
    \text{bpc}(\mathcal{S}) = \frac{\# \text{bits}(\mathcal{S})}{K}.
\end{equation}
By averaging over classes, bpc directly reports the per-class storage requirement in bits, irrespective of image format, label representation (hard vs. soft), or decoder complexity. Thus, bpc enables fair comparisons across DD methods in trading off rate and utility.

\section{Experiments}
\label{sec:Experiments}
To validate the effectiveness of our joint rate–utility optimization method, we conduct extensive evaluations on CIFAR-10, CIFAR-100, and an ImageNet subset. We assess classification accuracy under varying bpc budgets, test robustness across network architectures, and verify compatibility with different distillation losses.

\subsection{Experimental Setups}

\noindent\textbf{Datasets.} We evaluate on CIFAR-10 and CIFAR-100 at $32\times 32$ resolution, and on six subsets of ImageNet (at $128\times 128$ resolution), namely Nette, Woof, Fruit, Yellow, Meow, and Squawk~\cite{shin2025distilling}.

\noindent \textbf{Baselines.} Comparisons include TM~\cite{cazenavette2022dataset},  
GLaD~\cite{Cazenavette2023GeneralizingDD}, H-GLaD~\cite{zhong2024hierarchical}, FRePo~\cite{zhou2022dataset}, HaBa~\cite{liu2022dataset}, IDC~\cite{pmlr-v162-kim22c}, 
FreD~\cite{shin2023frequency}, SPEED~\cite{wei2023sparse}, RTP~\cite{deng2022remember}, NSD~\cite{yang2024neural}, HMN~\cite{zheng2024leveraging}, and DDiF~\cite{shin2025distilling}.

\noindent\textbf{Implementation Details.} We adopt the TM loss~\cite{cazenavette2022dataset} as our default utility objective. To accelerate convergence, all parameters are initialized via an ``overfitted'' image compression pretraining~\cite{kim2024c3} on the original dataset $\mathcal{D}$, with the following objective:
\begin{equation} \min_{\mathcal{Z},\Phi,\Psi}  r(\mathcal{Z};\Phi) + \beta\left\Vert g\left(z;\psi\right) -x \right\Vert_2^2,
\label{eq:init}
\end{equation}
where $z\in\mathcal{Z}$ is the latent code for input $x$, decoded by $g(\cdot;\psi)$ with $\psi \subset\Psi$. We then jointly optimize the entire method with Adam for the objective in Eq.~\eqref{eq:simob} until convergence. We evaluate our approach by training downstream classifiers across five independent trials and reporting the mean performance. Further implementation details are provided in the Appendix.

\begin{table}[t]
  \centering
  \caption{Classification accuracies (\%) on the $128 \times 128$ ImageNet subsets (\ie, Nette, Woof, Fruit, Yellow, Meow, and Squawk), evaluated under a bpc budget of $  \le 192$ kB.  The performance when trained on the original dataset is listed in the first row. A dash ``--'' denotes results not reported in the original work. }  
  \renewcommand\arraystretch{0.95}
  \setlength{\tabcolsep}{2.5pt}{ 
  \small
    \label{tab:bpc-imagesub}
  \begin{tabular}{lcccccccc}
  \toprule
    \multirow{2}{*}{Method} & \multicolumn{6}{c}{ImageNet Subset ($128\times128$)}  \\
  \cmidrule(lr){2-7} 
    &Nette & Woof & Fruit & Yellow & Meow & Squawk  \\
  \midrule
 Original& {87.4} & {67.0} & {63.9} & {84.4} & {66.7} & {87.5}  \\  \cmidrule(lr){1-7} 
  TM (Vanilla)~\cite{cazenavette2022dataset} & 51.4 & 29.7 & 28.8 & 47.5 & 33.3 &41.0 \\
  GLaD~\cite{Cazenavette2023GeneralizingDD} &38.7 &23.4 & 23.1 &-- & 26.0 &35.8\\
  H-GLaD~\cite{zhong2024hierarchical} & 45.4 & 28.3 & 25.6 & -- & 29.6 &39.7\\
  FRePo~\cite{zhou2022dataset} & 48.1 & 29.7 & -- & -- & -- & -- \\
  HaBa~\cite{liu2022dataset} & 51.9 & 32.4 & 34.7 & 50.4 & 36.9 & 41.9 \\
  IDC~\cite{pmlr-v162-kim22c} & 61.4 & 34.5 & 38.0 & 56.5 & 39.5 & 50.2  \\
  FreD~\cite{shin2023frequency} & 66.8 & 38.3 & 43.7 &63.2 &43.2 &57.0  \\
  SPEED~\cite{wei2023sparse} & 66.9 & 38.0 & 43.4 & 62.6 & 43.6 &60.9\\
  NSD~\cite{yang2024neural}  & 68.6 & 35.2 & 39.8 & 61.0 & 45.2 & 52.9 \\  
  DDiF~\cite{shin2025distilling} & 72.0 & 42.9 & 48.2 & 69.0 & 47.4 & 67.0  \\
  \hline
  TM-RUO (Ours)  &\textbf{76.5} & \textbf{46.0} & \textbf{50.8} & \textbf{74.3} & \textbf{53.5} & \textbf{74.8} \\
  \bottomrule
  \end{tabular}
  }
  \end{table}
  
\begin{table}[t]
\centering
\caption{Classification accuracies (\%) averaged across six $128 \times 128$ ImageNet subsets for different network architectures, including AlexNet, VGG-11, ResNet-18, and a variant of ViT-B, evaluated under a bpc budget of $
  \le 192$ kB.}
\label{tab:imagenet_subset_ipc1}
\renewcommand\arraystretch{0.95}
\setlength{\tabcolsep}{4pt}
\small
\begin{tabular}{lccccc}
\toprule
\multirow{2}{*}{Method} & \multicolumn{4}{c}{Classifier Architecture} & \multirow{2}{*}{Avg} \\
\cmidrule(lr){2-5}
 & AlexNet & VGG-11 & ResNet-18 & ViT \\
\midrule
TM (Vanilla)~\cite{cazenavette2022dataset} & 10.8 & 14.5 & 29.0 & 19.7 & 18.5 \\
IDC~\cite{pmlr-v162-kim22c} & 18.3 & 16.4 & 36.7 & 28.1 & 24.9 \\
FreD~\cite{shin2023frequency} & 24.3 & 16.9 & 40.0 & 32.9 & 28.5 \\
DDiF~\cite{shin2025distilling} & \textbf{49.3} & 38.6 & 49.6 & \textbf{43.5} & 45.2 \\
\hline
TM-RUO (Ours) & 46.7 & \textbf{56.7} & \textbf{56.0} & 43.4 & \textbf{50.7} \\
\bottomrule
\end{tabular}
\end{table}

\begin{table}[t]
\centering
\caption{Classification accuracies (\%) averaged across six $128 \times 128$ ImageNet subsets for different utility losses, including gradient matching (GM)~\cite{zhao2020dataset} and distribution matching (DM)~\cite{Zhao_2023_WACV}, evaluated under a bpc budget of  $\le 192$ kB.  An asterisk ``*'' indicates results reproduced by us.}
\label{tab:ddd}
\renewcommand\arraystretch{0.95}
\setlength{\tabcolsep}{2pt}{
\small
\begin{tabular}{lccccccc}
\toprule
\multirow{2}{*}{Method} & \multicolumn{6}{c}{ImageNet Subset ($128\times128$)} & \multirow{2}{*}{Avg} \\
\cmidrule(lr){2-7}
& Nette & Woof & Fruit& Yellow & Meow & Squawk \\
\midrule
\rowcolor{gray!30}
GM & & & & & & &\\
GM (Vanilla)~\cite{zhao2020dataset} & 34.2 & 22.5 & 21.0 &37.1\textsuperscript{*} & 22.0 &32.0 & 28.1 \\
GLaD~\cite{Cazenavette2023GeneralizingDD} & 35.4 & 22.3 & 20.7 &--& 22.6 & 33.8 & 27.0 \\
H-GLaD~\cite{zhong2024hierarchical} & 36.9 & 24.0 & 22.4 &--& 24.1 & 35.3 & 28.5 \\
IDC~\cite{pmlr-v162-kim22c} & 45.4 & 25.5 & 26.8 & -- &25.3 & 34.6 & 31.5 \\
 FreD~\cite{shin2023frequency} & 49.1 & 26.1 & 30.0 &--& 28.7 & 39.7 & 34.7 \\
DDiF~\cite{shin2025distilling} & 61.2 & \textbf{35.2} & \textbf{37.8} &--& 39.1 & 54.3 & 45.5 \\
\hline
GM-RUO (Ours) & \textbf{67.2} & 33.1 & 37.0 &\textbf{58.9}& \textbf{42.0} & \textbf{58.6} &\textbf{49.5}\\
\midrule
\rowcolor{gray!30}
DM & & & & & & &\\
DM (Vanilla)~\cite{Zhao_2023_WACV} & 30.4 & 20.7 & 20.4 &36.0\textsuperscript{*} & 20.1 & 26.6 & 25.7 \\
GLaD~\cite{Cazenavette2023GeneralizingDD} & 32.2 & 21.2 & 21.8&-- & 22.3 & 27.6 & 25.0 \\
H-GLaD~\cite{zhong2024hierarchical} & 34.8 & 23.9 & 24.4 & --&24.2 & 29.5 & 27.4 \\
IDC~\cite{pmlr-v162-kim22c} & 48.3 & 27.0 & 29.9 &-- &30.5 & 38.8 & 34.9 \\
FreD~\cite{shin2023frequency} & 56.2 & 31.0 & 33.4 &--& 33.3 & 42.7 & 39.3 \\
DDiF~\cite{shin2025distilling} & 69.2 & 42.0 & 45.3 &--& 45.8 & 64.6 & 53.4 \\
\hline
DM-RUO (Ours) & \textbf{71.9} & \textbf{46.4} & \textbf{49.0} &\textbf{69.2}& \textbf{48.8} & \textbf{69.0} &\textbf{59.1}\\
\bottomrule
\end{tabular}}
\end{table}

\subsection{Main Results}
\noindent \textbf{Results on CIFAR-10, CIFAR-100, and ImageNet Subsets.} 
Fig.~\ref{fig:RU} shows the rate-utility curves on CIFAR-10 and CIFAR-100. Our method, coupled with the TM loss and termed as TM-RUO, consistently outperforms all competing methods, achieving $77.5\%$ accuracy at $94$ kB, representing up to a $2.5\%$ absolute gain over the next best method. On CIFAR-100, TM-RUO attains $44.4 \%$ and $49.2\%$ accuracy at only $8$ kB and $53$ kB, respectively, setting new state-of-the-art trade-offs. Similar conclusions can be drawn on ImageNet subsets (see Fig.~\ref{fig:ru_nette} and Table \ref{tab:bpc-imagesub}).

\begin{figure*}[t]
\centering
\includegraphics[width=1\linewidth]{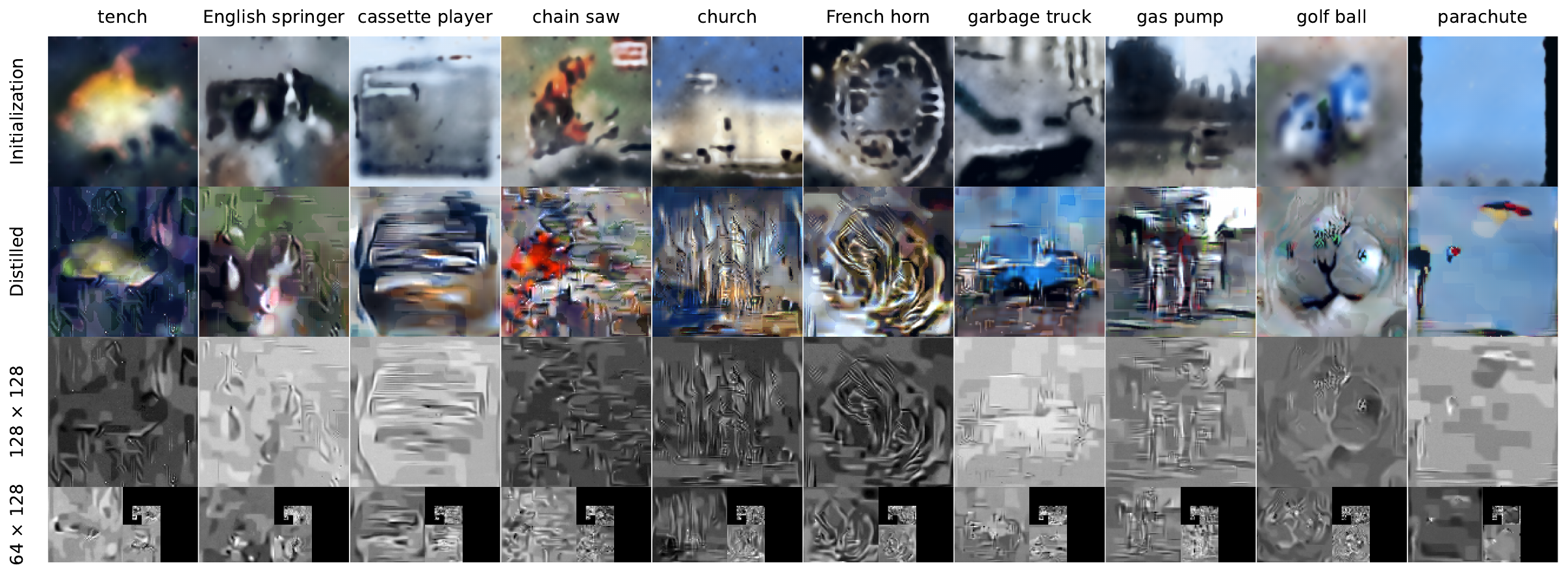}
\caption{Visualization of synthetic samples on the Nette subset of ImageNet. The first row presents initialization by ``overfitted'' image compression, the second row shows distilled images (from the post-quantized decoder), and the third and fourth rows display the corresponding multiscale latent codes. Zoom in for improved visibility.}
\label{fig:vis_syn}
\end{figure*}

\noindent \textbf{Cross-architecture Generalization.} 
To assess the generalizability of synthetic data across architectures, we train classifiers implemented by four widely used networks---AlexNet~\cite{krizhevsky2012imagenet}, VGG-11~\cite{simonyan2014very}, ResNet-18~\cite{he2016deep}, and a variant of ViT-B~\cite{DosovitskiyB0WZ21}---that differ from the network architecture employed during distillation. As shown in Table~\ref{tab:imagenet_subset_ipc1}, under a stringent bitrate budget (\ie, bpc $\le 192$ kB)\footnote{$\text{bpc}=192$ kB corresponds to $\text{ipc}=1$ of image dimensions  $128\times 128 \times 3$, where each pixel is represented at a $32$-bit depth.}, TM-RUO 
attains the highest accuracies across almost all architectures, with an average accuracy of $50.7\%$. These results demonstrate the robustness and adaptability of the proposed method in cross-architecture scenarios.

\noindent \textbf{Compatibility with Multiple DD Losses.} 
We also plug in gradient matching (GM)~\cite{zhao2020dataset} and distribution matching (DM)~\cite{Zhao_2023_WACV} losses into our method. As shown in Table~\ref{tab:ddd}, GM-RUO and DM-RUO achieve $49.5\%$ and $59.1\%$, respectively, both surpassing all competing methods and demonstrating broad compatibility.

\begin{figure}[t]
  \centering
  \includegraphics[width=0.99\linewidth]{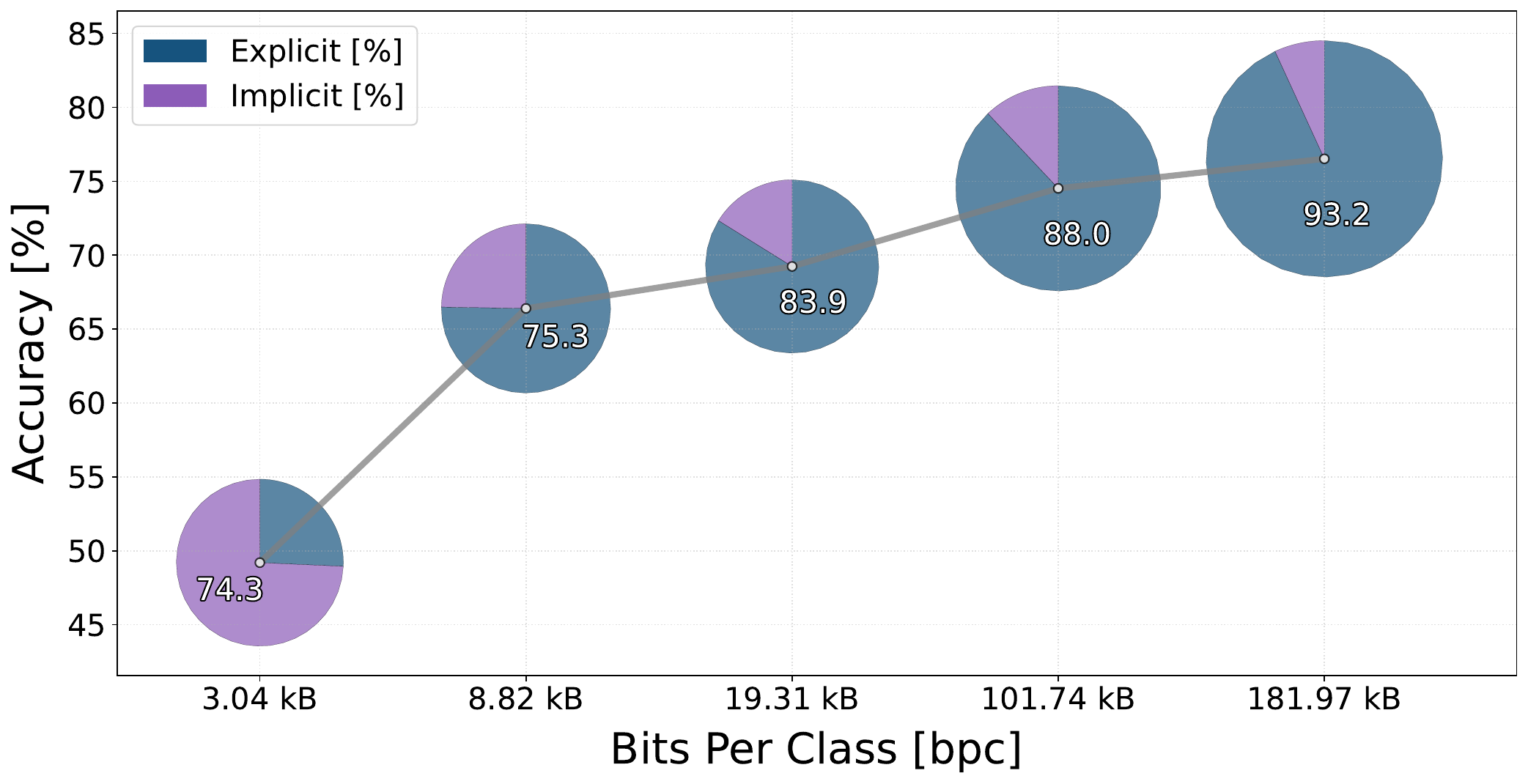}
  \caption{Bit allocation analysis across bpc regimes, with post-quantization error $<5\!\times\!10^{-7}$ in terms of mean squared error.}
  \label{fig:abla-bit}
  \end{figure}

\noindent \textbf{Synthetic Sample Visualization.} In Fig.~\ref{fig:vis_syn}, we visualize the optimized synthetic samples and their multiscale latent codes, alongside ``overfitted'' initialization from the Nette subset. It is clear that the explicit latents capture coarse-to-fine object contours, while the implicit synthesis injects category-specific details, validating the complementary roles of our hybrid representation.

\noindent \textbf{Bit Allocation Analysis.} 
Fig.~\ref{fig:abla-bit} examines how explicit and implicit bits distribute across bpc regimes. As our entropy and decoder networks are held constant, at low bpc (\eg, $3.04$ kB) implicit bits dominate ($\approx 74\%$), whereas at high bpc (\eg, $182$ kB), explicit latents account for $\approx 93\%$. This trend suggests that tight budgets require balancing explicit diversity and implicit structure, while generous budgets benefit more from explicit feature compression.

\section{Conclusion}
\label{sec:Conclusion}
In this work, we have reframed DD as a joint rate-utility optimization problem. By introducing a differentiable objective that incorporates the bitrate of latent codes, class labels, and decoder parameters alongside a task utility loss, we achieved state-of-the-art trade-offs on CIFAR-10, CIFAR-100, and ImageNet-128. Our multiscale latent representations, decoded via lightweight networks, not only decouple sample-specific details from shared network structures but also enable transparent comparison through a consistent bpc metric. Empirically, this formulation yields up to $170\times$ compression relative to vanilla distillation while maintaining---or even improving---test accuracy under tight storage budgets.

Looking ahead, we see three main avenues for extending this paradigm. First, fully joint optimization of soft labels and network architectures promises to tighten the rate-utility frontier: by learning continuous label embeddings rather than one-hot encodings, and by integrating differentiable neural architecture search to co-design entropy and decoder network topologies, one can further reduce bitrate without sacrificing accuracy. Second, scalability remains a practical bottleneck for high-resolution and large-scale datasets. Incorporating recent decoupling techniques, such as staged pipelines like SRe2L~\cite{yin2023squeeze}, will drastically cut computation and memory costs.

Finally, we believe our joint rate-utility optimization method can be generalized far beyond images. Extending to video, text, and graph modalities will require spatiotemporal or relational decoder architectures and entropy models attuned to temporal and structural dependencies. Moreover, adaptive schemes---where bit allocations and loss weights vary per sample, class, or downstream task---could enable conditional or multi-task distillation, dynamically trading storage for performance where it matters most. By pursuing these directions---soft label compression, neural architecture search, efficiency-focused decoupling tricks, and richer compression tools---we anticipate that rate-utility optimization will become a versatile foundation for data-efficient learning across domains.

\section*{Acknowledgments}
This work was supported in part by the National Natural Science Foundation of China (62472124), Hong Kong ITC Innovation and Technology Fund (9440379 and 9440390), Shenzhen Colleges and Universities Stable Support Program (GXWD20220811170130002), and Major Project of Guangdong Basic and Applied Basic Research (2023B0303000010).

{
    \small
    \bibliographystyle{ieeenat_fullname}
    \bibliography{main}
}


\counterwithin{figure}{section}  
\counterwithin{table}{section}   
\counterwithin{equation}{section} 
\clearpage
\setcounter{page}{1}

\onecolumn
\section*{Appendix}
This appendix elaborates on 1) the information-theoretic cost of encoding soft labels, 2) full hyperparameter and architecture configurations in our joint rate-utility optimization method, 3) expanded quantitative results, 4) runtime statistics for synthetic dataset generation, and 5) more synthetic data visualizations.
\renewcommand{\thesubsection}{A\arabic{subsection}} 

\renewcommand{\thefigure}{A\arabic{figure}}    
\renewcommand{\thetable}{A\arabic{table}}      
\renewcommand{\theequation}{A\arabic{equation}}

\setcounter{figure}{0}
\setcounter{table}{0}
\setcounter{equation}{0}
\subsection{Bitrate for Soft Labels}\label{subsec:bsl}
Let $\mathcal{Y} = \{\tilde{y}^{(i)}\}_{i=1}^N$ be a collection of soft classes, where each  $\tilde{y}^{(i)}$ is a probability vector
\begin{equation}
   \tilde{y}^{(i)} = [\tilde{y}^{(i)}_1, \ldots, \tilde{y}^{(i)}_K], \quad \tilde{y}^{(i)}_k \ge 0, \quad \sum_{k=1}^K \tilde{y}^{(i)}_k = 1,
\end{equation}
lying in the $(K-1)$-simplex $\Delta^{K-1}$. To encode these continuous vectors, we quantize $\Delta^{K-1}$ into $B$ cells (\ie, ``bins'') of side length $\epsilon$. Denote by $\{\Omega_i\}_{i=1}^B$ the partition cells, each with volume $\mathrm{Vol}(\Omega_i)\approx \epsilon^{K-1}$, and let 
\begin{equation}
    P_i = \int_{\Omega_i} p(\xi)\mathrm{d}\xi
\end{equation}
be the probability mass of soft labels falling in bin $\Omega_i$ under the true (non-uniform) density $p(\xi)$. The optimal bitrate per label, using entropy coding, is given by the Shannon entropy of the mass distribution:
\begin{equation}
    r(\mathcal{Y}) \approx \mathrm{Entropy}(P) = -\sum_{i=1}^B P_i \log_2 P_i,
\end{equation}
which accounts for non-uniform concentrations of soft labels across the simplex. As entropy is maximized when $P_i = 1/B$ for all $i$, we have
\begin{align}
    r(\mathcal{Y}) \leq \log_2 B \approx& \log_2\left(\frac{\mathrm{Vol}(\Delta^{K-1})}{\epsilon^{K-1}}\right)\nonumber\\
    =& \log_2\left(\frac{1}{(K-1)!\epsilon^{K-1}}\right)\nonumber\\
    =&  -\log_2(K-1)!- (K-1)\log_2\epsilon.
\end{align}
where $\mathrm{Vol}(\Delta^{K-1}) = 1/(K-1)!$. This recovers the bound in the main text for worst-case (uniform) quantization.

\noindent\textbf{Practical illustration.} For single-precision floats (\ie, \texttt{fp32}) the $24$-bit mantissa roughly yields a machine precision of $\epsilon\approx 2^{-24}$. For a $K=1,000$-class problem such as ImageNet-1K classification, the continuous label space $\mathcal{Y}$ incurs an encoding cost of $r(\mathcal{Y}) \approx 15,456$ bits/vector, which dwarfs the $\log_2 K = \log_2 1,000\approx 10$ bits needed for representing a hard (one-hot) label, even surpassing the size of synthetic samples themselves.

\subsection{Hyperparameter Settings}
\noindent\textbf{Implementation Details.}
For synthetic dataset parameterization, we employ the C3 defaults~\cite{kim2024c3}: $L=6$ latent scales with an upsampling kernel of size $8$. We customize both the entropy and decoder networks to accommodate different datasets and varying synthetic samples per class (spc). The entropy network is implemented by a multilayer perceptron, whose detailed layer-by-layer specification is provided in Table~\ref{tab:hyper_all}. All hidden layers employ ReLU activations, and the final layer outputs two values---$\mu, \log\sigma$---which parameterize the conditional  Laplace distribution of each latent code. To enforce causality, latent codes are processed in raster-scan order~\cite{OordKK16}. For each code $m$ at scale $l$, we extract its causal context from a fixed-size neighborhood of $C$ previously encoded codes, experimenting with  $C \in \{8, 16, 24, 32, 64\}$ (see Table~\ref{tab:hyper_all} for more details).
A single entropy network then processes all scales simultaneously by concatenating context tensors  $\{c_1, \ldots, c_{L}\}$, where ${c}_l \in \mathbb{R}^{\left(\left\lfloor\frac{H}{2^{l-1}}\right\rfloor \times\left\lfloor\frac{W}{2^{l-1}}\right\rfloor\right) \times C}$, to form $c \in\mathbb{R}^{\left(\sum_{l=1}^L\left\lfloor\frac{H}{2^{l-1}}\right \rfloor\times\left\lfloor\frac{W}{2^{l-1}}\right\rfloor\right) \times C}$.

Fig.~\ref{fig:our_framework} in the main text depicts our decoder architecture, which comprises five convolution layers with hyperparameters $\{D_1, D_2, D_3\}$. Two residual connections link the last two convolutions. Following the guidelines of~\cite{kim2024c3}, we implement eight distinct decoder configurations (see Table~\ref{tab:network_config}), where setting $D_2=0$ indicates the identity mapping (\ie, no learnable parameters) at that layer. ReLU is adopted as the nonlinear activation function. 
We further denote ``slice size'' as the number of synthetic samples handled by each entropy or decoder network. 
By default, we allocate one network per class. However, for CIFAR-10 with $\text{spc}= 718$, we set the slice size to $359$, resulting in two slices per class.

For downstream classifiers, we follow FreD~\cite{shin2023frequency} and employ convolutional neural networks for both dataset distillation and evaluation. The architecture comprises identical blocks, each containing a $3\times 3$ convolution with $128$ filters, instance normalization, ReLU, and $2\times 2$ average pooling (stride $2$). A final linear layer produces the output logits. We configure three blocks for $32\times32$ datasets (\ie, CIFAR‑10 and CIFAR‑100), and five blocks for $128\times128$ ImageNet subsets.

We use TM~\cite{cazenavette2022dataset} as our default distillation loss, while supporting GM~\cite{zhao2020dataset} and DM~\cite{Zhao_2023_WACV} identically. All experiments use differentiable siamese augmentation~\cite{zhao2022synthesizing} for data augmentation; zero-phase component analysis as a form of signal whitening is only applied to CIFAR-10 at $\text{spc}= 64$ or $240$. Hyperparameters for each loss generally follow those of FreD~\cite{shin2023frequency} and DDiF~\cite{shin2025distilling} (see Table~\ref{tab:hyper_all} for details), with synthetic minibatch sizes adjusted for higher spc in TM.

\noindent\textbf{Training Protocols.} In the \textit{initialization} phase, we adopt the default warm-up schedule from C3~\cite{kim2024c3}, setting the learning rate to $0.01$. The rate-distortion trade-off coefficient $\beta$ is chosen per dataset and distillation loss. For instance, 
 $\beta = 10$ when applying the TM loss on ImageNet subsets, while $\beta = 10^{6}$ for both the GM and DM losses.

In the \textit{joint rate-utility optimization} phase, training is carried out using the Adam optimizer at a fixed learning rate of $10^{-3}$. We run $15,000$ iterations for TM, $800$ for GM, and $20,000$ for DM, respectively. 
Bit-budget enforcement is governed by a two-stage schedule for the rate-utility coefficient $\lambda$: during the first half of training, a generally higher $\lambda$ prioritizes distillation performance; in the second half, we are inclined to decrease $\lambda$
to strictly impose the bitrate constraint. 
In the \textit{post-quantization} phase, we enforce the mean squared error between the pre- and post-quantized synthetic samples to lie within $\{5 \times 10^{-5}$, $5 \times 10^{-6}$, $5 \times 10^{-7}$, $5 \times 10^{-8}\}$, then select the threshold that maximizes rate-utility performance for the target bpc budget.
In the \textit{evaluation} phase, we train five independent classifiers for $1,000$ epochs and report the average classification accuracy. For cross-architecture comparisons, we utilize FreD’s implementations~\cite{shin2023frequency}, applying a fixed Adam learning rate of $0.02$ for AlexNet, VGG-11, ResNet-18, and $0.015$ (with a dropout ratio of $0.01$) for a variant of ViT-B. All experiments are performed on NVIDIA A100 ($4\times$) and A800 ($4\times$) GPUs.

\begin{table}[t]
\centering
\caption{Configuration of the decoder network across model variants, denoted by ``$\text{v}\langle\text{depth}\rangle\text{-}\langle \text{channels}\rangle$,'' where the suffix denotes the network depth and first‐layer output channels. $D_1$ to $D_3$ list the output channel counts for layers one through three, and the last column gives the total number of trainable parameters.}
\label{tab:network_config}
\renewcommand\arraystretch{1}
\setlength{\tabcolsep}{15pt}{
\begin{tabular}{lcccc}
\toprule
 Version & $D_{1}$ & $D_{2}$ &$D_{3}$ & $\#$Parameters\\
\midrule
v4-40    & 40   & 0    & 3 & 571 \\ 
v4-160   & 160  & 0    & 3 & 1,771 \\ 
v4-240   & 240  & 0    & 3 & 2,571 \\ 
v4-480   & 480  & 0    & 3 & 4,971 \\ 
v4-960   & 960  & 0    & 3 & 9,771 \\ 
v4-1200  & 1,200 & 0    & 3 & 12,171 \\ 
v5-240   & 240  & 40   & 3 & 11,611 \\ 
v5-320   & 320  & 40   & 3 & 15,371 \\
\bottomrule
\end{tabular}
}
\end{table}

\begin{table*}[t]
\centering
\caption{Overview of hyperparameter settings for different DD loss functions. }
\label{tab:hyper_all}
\begin{subtable}{1.0\textwidth}
\centering
\caption{Hyperparameter settings for the TM loss.}
\setlength{\tabcolsep}{4.5pt}
{
\begin{tabular}{ll*{10}{c}} 
  \toprule
  \multirow{3}{*}{Dataset}
  &\multirow{3}{*}{spc} 
  & \multicolumn{4}{c}{TM} 
  & \multicolumn{2}{c}{Entropy Network} 
  & \multicolumn{2}{c}{Decoder Network}
  & \multicolumn{2}{c}{Optimization} \\ 
  \cmidrule(r){3-6} \cmidrule(r){7-8} \cmidrule(r){9-10} \cmidrule(r){11-12} 
  &
  & \makecell{$t_{1}$} 
  & \makecell{$t_{2}$} 
  & \makecell{$t$} 
  & \makecell{Synthetic\\ minibatch size} 
  & \makecell{Width} 
  & \makecell{Depth} 
  & \makecell{Version} 
  & \makecell{Slice Size}
  & \makecell{$\beta$} 
  & \makecell{$\lambda$} \\
  \midrule
  \multirow{3}{*}{CIFAR-10} 
  &64  & 60 & 392 & 1,960 & 256  & 16 & 4 & v4-480 & 64& $10^{6}$ & {$\{2\times10^{1}\}$}  \\
  &240 & 60 & 392 & 7,840 & 410   & 16 & 4 & v4-960 & 240&$10^{6}$ & {$\{10^{2}, 2\times10^{2}\}$} \\
  &718 & 60 & 392 & 7,840 & 720  & 16 & 4 & v4-1200 & 360&$10^{6}$ & $\{2\times10^{2}\}$ \\
  \midrule
  \multirow{2}{*}{CIFAR-100} 
  &48  & 60 & 392 & 7,840 & 256   & 16 & 4 & v4-240 &48& $10^{8}$ & {$\{3\times10^{2}\}$} \\
  &120 & 40 & 392 & 9,408 & 960 & 16 & 4 & v4-960 &120& $10^{6}$ & {\{$1.5\times10^{3}\}$} \\
  \midrule
  \multirow{5}{*}{\makecell{ImageNet Subset}}
  &1   & 20 & 102 & 510 & 102  & 16 & 2 & v4-40 &1& $10$ & {$\{10^{2}, 10^{3}\}$}\\
  &8   & 20 & 102 & 510 & 102 & 16 & 2 & v4-160 & 8&$10$ & {$\{10^{3}, 8.5\times10\}$}\\
  &15  & 20 & 102 & 510 & 102  & 16 & 2 & v4-240 & 15&$10$ & {$\{10^{3}, 8.5\times10\}$} \\
  &51  & 40 & 102 & 1,020 & 80   & 32 & 4 & v5-240 & 51&$10$ & {$\{10^{3}, 8.5\times10\}$}\\
  &102 & 40 & 102 & 1,020 & 80   & 32 & 4 & v5-240 & 102&$10$ & {$\{10^{3}, 6.6\times10\}$}\\
  \bottomrule
\end{tabular}
}
\end{subtable}

\begin{subtable}{\textwidth}
\centering
\caption{Hyperparameter settings for the GM and DM losses.}
\setlength{\tabcolsep}{6pt}
\begin{tabular}{lccccccccc}
    \toprule
    \multirow{3}{*}{Dataset} 
    &\multirow{3}{*}{$\ell$} 
    & \multirow{3}{*}{spc} 
    &\multirow{3}{*}{\makecell{Synthetic\\ batch size} } 
& \multicolumn{2}{c}{Entropy Network} 
  & \multicolumn{2}{c}{Decoder Network}
    & \multicolumn{2}{c}{Optimization} \\
   \cmidrule(lr){5-6} \cmidrule(lr){7-8} \cmidrule(lr){9-10}
    & 
    & &
    & \makecell{Width} 
    & \makecell{Depth} 
    & \makecell{Version} 
    & \makecell{Slice Size}
    & \makecell{$\beta$} 
    & \makecell{$\lambda$} \\
    \midrule
    \multirow{2}{*}{\makecell{ImageNet Subset}} 
    & GM
    & 96 
    & 960
    & 32 & 4 & v5-320 &96
    & $10^{6}$ & {$\{2.8 \times 10^{-2}, 1.2 \times 10^{-2}\}$} \\
    & DM
    & 96
    & 960
    & 32 & 4 & v5-320 &96
    & $10^{6}$ & {$\{2 \times 10^{1}, 6.7 \times 10^{-1}\}$} \\
    \bottomrule
\end{tabular}
\end{subtable}
\end{table*}

\subsection{Expanded Quantitative Results}
We provide detailed performance comparisons across multiple datasets and distillation losses to underscore the generality and efficiency of our joint rate-utility optimization method.

Table~\ref{tab:nette} supplements the raw data for the rate-utility curve in Fig.~\ref{fig:ru_nette} of the main text.

Table \ref{tab:cifar10_results} 
summarizes results on CIFAR-10 and CIFAR-100 over a range of bpc budgets. Under each constraint, our method not only achieves state-of-the-art classification accuracy but also requires substantially fewer bits---for example, on CIFAR-100 with a $120$ kB budget, we attain superior accuracy using only $53$ kB.

Table~\ref{tab:my_table_5} extends this analysis by applying the TM loss~\cite{cazenavette2022dataset}  across four distinct backbone architectures (\ie, AlexNet, VGG-11, ResNet-18, and a variant of ViT-B), confirming cross-architecture generalization.

Table~\ref{tab:my_table_4} reports averaged accuracies ($\pm$ standard deviation) on the ImageNet subsets, when using the GM~\cite{zhao2020dataset} and DM~\cite{Zhao_2023_WACV} distillation losses under $\text{bpc}  = 192$ kB with $\text{spc} = 96$, revealing consistent gains over vanilla baselines.

\subsection{Wall-Clock Time}
We evaluate decoding speed on a single NVIDIA A100 80 GB GPU, averaged across $10,000$ runs. 
From Table~\ref{tab:decoding_time}, we find that total latency grows roughly linearly with spc, but the slope varies by dataset due to the differences in the entropy and decoder networks. For instance, increasing spc from $64$ to $718$ on CIFAR-10 raises average latency from $55.30$ ms to $318.57$ ms ($5.8\times$), while on ImageNet an increase from $1$ to $102$ spc yields a $5.7\times$ rise ($51.95$ ms to $295.87$ ms). Notably, per-sample latency improvements taper off at high spc: beyond $\text{spc}= 51$, ImageNet’s per-sample time only drops from $0.32$ ms to $0.29$ ms, and CIFAR-10 stabilizes at $0.04$ ms past $\text{spc}= 240$, indicating hardware or algorithmic limits on batching efficiency.

\begin{table*}[t]
\centering
\caption{Detailed experimental results on the Nette subset of ImageNet $(128\times128)$. Classification accuracies (\%) are reported as mean~$\pm$~standard deviation.}
\label{tab:nette}
\renewcommand\arraystretch{1}
\setlength{\tabcolsep}{20pt}
{
\begin{tabular}{lcccc}
\toprule
Method & spc & $\#$Parameters Per Class ($\times 10^4$) & bpc (kB) & Accuracy (\%) \\
\midrule
\multirow{4}{*}{TM~\cite{cazenavette2022dataset}} 
&  1 & 4.92 &  192.0 & 51.4{\scriptsize$\pm$2.3} \\
&  10 & 49.15 & 1,920.0 & 63.0{\scriptsize$\pm$1.3} \\
&  50 & 245.76 & 9,600.0 & 72.8{\scriptsize$\pm$0.8}  \\
&  51 & 250.68 & 9,792.0 & 73.0{\scriptsize$\pm$0.7}  \\
\midrule
GLaD~\cite{Cazenavette2023GeneralizingDD}  
& -- & 4.92 & 192.0 & 38.7{\scriptsize$\pm$1.6} \\
\midrule
H-GLaD~\cite{zhong2024hierarchical}  
& -- & 4.92 & 192.0 & 45.4{\scriptsize$\pm$1.1}  \\
\midrule
\multirow{2}{*}{FRePo~\cite{zhou2022dataset}} 
& -- & 4.92 & 192.0 & 48.1{\scriptsize$\pm$0.7}  \\
& -- & 49.15 & 1,920.0 & 66.5{\scriptsize$\pm$0.8}  \\
\midrule
\multirow{2}{*}{HaBa~\cite{liu2022dataset}} 
& -- & 4.92 & 192.0 & 51.9{\scriptsize$\pm$0.7}  \\
& -- & 49.15 & 1,920.0 & 64.7{\scriptsize$\pm$1.6}  \\
\midrule
\multirow{2}{*}{IDC~\cite{pmlr-v162-kim22c}} 
& -- & 4.92 & 192.0 & 61.4{\scriptsize$\pm$1.0} \\
& -- & 49.15 & 1,920.0 & 70.8{\scriptsize$\pm$0.5}  \\
\midrule
\multirow{3}{*}{FReD~\cite{shin2023frequency}} 
&  8 & 4.92 & 192.0 & 66.8{\scriptsize$\pm$0.4}  \\
&  16 & 9.83 & 384.0 & 69.0{\scriptsize$\pm$0.9}  \\
&  40 & 49.15 & 1,920.0 & 72.0{\scriptsize$\pm$0.8}  \\
\midrule
\multirow{2}{*}{SPEED~\cite{wei2023sparse}} 
&  15 & 4.92 & 192.0 & 66.9{\scriptsize$\pm$0.7}  \\
& 111 & 49.15 & 1,920.0 & 72.9{\scriptsize$\pm$1.5}  \\
\midrule
NSD~\cite{yang2024neural} 
& -- & 4.92 & 192.0 & 68.6{\scriptsize$\pm$0.8}  \\
\midrule
\multirow{6}{*}{DDiF~\cite{shin2025distilling}} 
&  1 & 0.10 & 3.8 & 49.1{\scriptsize$\pm$2.0} \\
& 8  & 0.77 & 30.1 & 67.1{\scriptsize$\pm$0.4}  \\
& 15 & 1.44 & 56.4 & 68.3{\scriptsize$\pm$1.1}  \\
& 51 & 4.92 & 192.0 & 72.0{\scriptsize$\pm$0.9}  \\
& 510 & 49.15 & 1,920.0 & 74.6{\scriptsize$\pm$0.7}  \\
& 697 & 245.55 & 9,591.2 & 75.2{\scriptsize$\pm$1.3}  \\
\midrule
\multirow{5}{*}{{TM-RUO (Ours)}} 
& 1 & 2.31 & 3.2 & {49.2}{\scriptsize$\pm$1.2}  \\
& 8 & 17.71 & 8.4 & 66.4{\scriptsize$\pm$1.6}  \\
& 15 & 33.08 & 18.2 & {69.2}{\scriptsize$\pm$1.5}  \\
& 51 & 112.98 & 101.7 & {74.5}{\scriptsize$\pm$0.8}  \\
& 102 & 224.36 & 179.7 & {76.5}{\scriptsize$\pm$0.7}  \\
\bottomrule
\end{tabular}}
\end{table*}

\begin{table}[h]
  \centering
  \caption{Classification accuracies (mean $\pm$ standard deviation) on CIFAR-10 and CIFAR-100 under five different bpc budgets. The first row (``Original'') shows accuracy when training on the original dataset. The budget rows give the compressed-size budgets (in kB) with the corresponding number of synthetic spc  (in parentheses).}
  \label{tab:cifar10_results}
  \renewcommand\arraystretch{1}
  \setlength{\tabcolsep}{15pt}
  \begin{tabular}{lccccccccc}
  \toprule
  \multirow{2}{*}{Method} & \multicolumn{3}{c}{CIFAR-10} & \multicolumn{2}{c}{CIFAR-100} \\
  \cmidrule(lr){2-4} \cmidrule(lr){5-6}
   & 12 kB & 120 kB & 600 kB & 12 kB & 120 kB\\
 \midrule
 Original & \multicolumn{3}{c}{84.8{\scriptsize$\pm$0.1}} & \multicolumn{2}{c}{56.2{\scriptsize$\pm$0.3}}\\
 \midrule
TM~\cite{cazenavette2022dataset} & 46.3\scriptsize±0.8 & 65.3\scriptsize±0.7 & 71.6\scriptsize±0.2 & 24.3\scriptsize±0.3 & 40.1\scriptsize±0.4 \\
FRePo~\cite{zhou2022dataset} & 46.8\scriptsize±0.7 & 65.5\scriptsize±0.4 & 71.7\scriptsize±0.2 & 28.7\scriptsize±0.1 & 42.5\scriptsize±0.2 \\
IDC~\cite{pmlr-v162-kim22c} & 50.0\scriptsize±0.4 & 67.5\scriptsize±0.5 & 74.5\scriptsize±0.2 & -- & -- \\ 
HaBa~\cite{liu2022dataset} & 48.3\scriptsize±0.8 & 69.9\scriptsize±0.4 & 74.0\scriptsize±0.2 & -- & -- \\
FreD~\cite{shin2023frequency} & 60.6\scriptsize±0.8 & 70.3\scriptsize±0.3 & 75.8\scriptsize±0.1 & 34.6\scriptsize±0.4 & 42.7\scriptsize±0.2 \\
RTP~\cite{deng2022remember} & 66.4\scriptsize±0.4 & 71.2\scriptsize±0.4 & 73.6\scriptsize±0.5 & 34.0\scriptsize±0.4 & 42.9\scriptsize±0.7 \\
NSD~\cite{yang2024neural} & \underline{68.5}{\scriptsize±0.8} & 73.4\scriptsize±0.2 & 75.2\scriptsize±0.6 & 36.5\scriptsize±0.3 & \underline{46.1}{\scriptsize±0.2} \\
SPEED~\cite{wei2023sparse} & 63.2\scriptsize±0.1 & 73.5\scriptsize±0.2 & \underline{77.7}{\scriptsize±0.4} & 40.4\scriptsize±0.4 & 45.9\scriptsize±0.3 \\
HMN~\cite{zheng2024leveraging} & 65.7\scriptsize±0.3 & 73.7\scriptsize±0.1 & 76.9\scriptsize±0.2 & 36.3\scriptsize±0.2 & 45.4\scriptsize±0.2 \\
DDiF~\cite{shin2025distilling} & 66.5\scriptsize±0.4 & \underline{74.0}{\scriptsize±0.4} & 77.5\scriptsize±0.3 & \underline{42.1}{\scriptsize±0.2} & 46.0\scriptsize±0.2 \\
\midrule
\multirow{2}{*}{Method} & \multicolumn{3}{c}{CIFAR-10} & \multicolumn{2}{c}{CIFAR-100} \\
\cmidrule(lr){2-4} \cmidrule(lr){5-6}
 & 13 kB (64) & 94 kB (240) & 246 kB (718) & 8 kB (48) & 53 kB (120) \\   \midrule
{TM-RUO (Ours)}  & \textbf{70.3}{\scriptsize±0.7} & \textbf{77.5}{\scriptsize±0.3} & \textbf{79.7}{\scriptsize±0.3} & \textbf{44.4}{\scriptsize±0.3} & \textbf{49.2}{\scriptsize±0.4} \\
  \bottomrule
  \end{tabular}
\end{table}

\begin{table*}[ht]
\centering
\caption{Classification accuracies (mean ± standard deviation) on six $128 \times 128$ ImageNet subsets across different network architectures, including AlexNet, VGG-11, ResNet-18, and a variant of ViT-B, evaluated under a bpc budget of $ \le 192$ kB.}
\label{tab:my_table_5}
\renewcommand\arraystretch{1.0}
\setlength{\tabcolsep}{10pt}{
\begin{tabular}{llcccccc}
\toprule
{Classifier} & {Method} & {Nette} & {Woof} & {Fruit} & {Yellow} & {Meow} & {Squawk} \\
\midrule
\multirow{5}{*}{AlexNet} 
& TM~\cite{cazenavette2022dataset} & 13.2{\scriptsize$\pm$0.6} & 10.0{\scriptsize$\pm$0.0} & 10.0{\scriptsize$\pm$0.0} & 11.0{\scriptsize$\pm$0.2} & 9.8{\scriptsize$\pm$0.0} & -- \\
& IDC~\cite{pmlr-v162-kim22c} & 17.4{\scriptsize$\pm$0.9} & 16.5{\scriptsize$\pm$0.7} & 17.9{\scriptsize$\pm$0.7} & 20.6{\scriptsize$\pm$0.9} & 16.8{\scriptsize$\pm$0.5} & 20.7{\scriptsize$\pm$1.0} \\
& FreD~\cite{shin2023frequency} & 35.7{\scriptsize$\pm$0.4} & 23.9{\scriptsize$\pm$0.7} & 15.8{\scriptsize$\pm$0.7} & 19.8{\scriptsize$\pm$1.2} & 14.4{\scriptsize$\pm$0.5} & 36.3{\scriptsize$\pm$0.3} \\
& DDiF~\cite{shin2025distilling} & \underline{60.7}{\scriptsize$\pm$2.3} & \textbf{36.4}{\scriptsize$\pm$2.3} & \textbf{41.8}{\scriptsize$\pm$0.6} & \textbf{56.2}{\scriptsize$\pm$0.8} & \textbf{40.3}{\scriptsize$\pm$1.9} & \textbf{60.5}{\scriptsize$\pm$0.4} \\
\cline{2-8}
& TM-RUO (Ours) & \textbf{64.1}{\scriptsize$\pm$1.3} & \underline{32.6}{\scriptsize$\pm$1.8} & \underline{40.7}{\scriptsize$\pm$0.5} & \underline{49.9}{\scriptsize$\pm$2.4} & \underline{36.9}{\scriptsize$\pm$1.6} & \underline{55.9}{\scriptsize$\pm$2.5} \\
\midrule
\multirow{5}{*}{VGG-11} 
& TM~\cite{cazenavette2022dataset} & 17.4{\scriptsize$\pm$2.1} & 12.6{\scriptsize$\pm$1.8} & 11.8{\scriptsize$\pm$1.0} & 16.9{\scriptsize$\pm$1.1} & 13.8{\scriptsize$\pm$1.3} & -- \\
& IDC~\cite{pmlr-v162-kim22c} & 19.6{\scriptsize$\pm$1.5} & 16.0{\scriptsize$\pm$2.1} & 13.8{\scriptsize$\pm$1.3} & 16.8{\scriptsize$\pm$3.5} & 13.1{\scriptsize$\pm$2.0} & 19.1{\scriptsize$\pm$1.2} \\
& FreD~\cite{shin2023frequency} & 21.8{\scriptsize$\pm$2.9} & 17.1{\scriptsize$\pm$1.7} & 12.6{\scriptsize$\pm$2.6} & 18.2{\scriptsize$\pm$1.1} & 13.2{\scriptsize$\pm$1.9} & 18.6{\scriptsize$\pm$2.3} \\
& DDiF~\cite{shin2025distilling} & \underline{53.6}{\scriptsize$\pm$1.5} & \underline{29.9}{\scriptsize$\pm$1.9} & \underline{33.8}{\scriptsize$\pm$1.9} & \underline{44.2}{\scriptsize$\pm$1.7} & \underline{32.0}{\scriptsize$\pm$1.8} & \underline{37.9}{\scriptsize$\pm$1.5} \\
\cline{2-8}
&TM-RUO (Ours) & \textbf{70.3}{\scriptsize$\pm$2.5} & \textbf{42.1}{\scriptsize$\pm$2.9} & \textbf{44.6}{\scriptsize$\pm$1.4} & \textbf{66.9}{\scriptsize$\pm$0.4} & \textbf{49.6}{\scriptsize$\pm$1.4} & \textbf{66.8}{\scriptsize$\pm$2.8} \\

\midrule
\multirow{5}{*}{ResNet-18} 
& TM~\cite{cazenavette2022dataset} & 34.9{\scriptsize$\pm$2.3} & 20.7{\scriptsize$\pm$1.0} & 23.1{\scriptsize$\pm$1.5} & 43.4{\scriptsize$\pm$1.1}& 22.8{\scriptsize$\pm$2.2} & -- \\

&IDC~\cite{pmlr-v162-kim22c}
&43.6\scriptsize{$\pm1.3$}&23.2\scriptsize{$\pm0.8$}&32.9\scriptsize{$\pm2.8$}&44.2\scriptsize{$\pm3.5$}&28.2\scriptsize{$\pm0.5$}&47.8\scriptsize{$\pm1.9$}\\
&FreD~\cite{shin2023frequency}&48.8\scriptsize{$\pm1.8$}&28.4\scriptsize{$\pm0.6$}&34.0\scriptsize{$\pm1.9$}&49.3\scriptsize{$\pm1.1$}&29.0\scriptsize{$\pm1.8$}&50.2\scriptsize{$\pm0.8$}\\
& DDiF~\cite{shin2025distilling} & \underline{63.8}{\scriptsize$\pm$1.8} & \underline{37.5}{\scriptsize$\pm$1.9} & \underline{42.0}{\scriptsize$\pm$1.9} & \underline{55.9}\scriptsize{$\pm1.0$} & \underline{35.8}{\scriptsize$\pm$1.8} & \underline{62.6}{\scriptsize$\pm$1.5} \\
\cline{2-8}
& TM-RUO (Ours)& \textbf{70.6}{\scriptsize$\pm$1.2} & \textbf{39.9}{\scriptsize$\pm$1.8} & \textbf{44.6}{\scriptsize$\pm$0.6} & \textbf{67.1}{\scriptsize$\pm$1.5} & \textbf{47.8}{\scriptsize$\pm$0.9} & \textbf{66.1}{\scriptsize$\pm$1.6} \\
\midrule
\multirow{5}{*}{Variant of ViT-B} 
& TM~\cite{cazenavette2022dataset} & 22.6{\scriptsize$\pm$1.1} & 15.9{\scriptsize$\pm$0.4} & 23.3{\scriptsize$\pm$0.4} & 18.1{\scriptsize$\pm$1.3} & 18.6{\scriptsize$\pm$0.9} & -- \\
& IDC~\cite{pmlr-v162-kim22c} & 31.0{\scriptsize$\pm$0.6} & 22.4{\scriptsize$\pm$0.8} & 31.1{\scriptsize$\pm$0.8} & 30.3{\scriptsize$\pm$0.6} & 21.4{\scriptsize$\pm$0.7} & 32.2{\scriptsize$\pm$1.2} \\
& FreD~\cite{shin2023frequency} & 38.4{\scriptsize$\pm$0.7} & 25.4{\scriptsize$\pm$1.7} & 31.9{\scriptsize$\pm$1.4} & 37.6{\scriptsize$\pm$2.0} & 19.7{\scriptsize$\pm$0.8} & 44.4{\scriptsize$\pm$1.0} \\
& DDiF~\cite{shin2025distilling} & \textbf{59.0}{\scriptsize$\pm$0.4} & \textbf{32.8}{\scriptsize$\pm$0.8} & \underline{39.4}{\scriptsize$\pm$0.8} & \underline{47.9}{\scriptsize$\pm$0.6} & \underline{27.0}{\scriptsize$\pm$0.6} & \underline{54.8}{\scriptsize$\pm$1.1} \\
\cline{2-8}
& TM-RUO (Ours) & \underline{57.5}{\scriptsize$\pm$1.2} & \underline{29.5}{\scriptsize$\pm$0.3} & \textbf{41.4}{\scriptsize$\pm$1.1} & \textbf{48.2}{\scriptsize$\pm$0.8} & \textbf{28.0}{\scriptsize$\pm$0.9} & \textbf{55.8}{\scriptsize$\pm$1.4} \\
\bottomrule
\end{tabular}}
\end{table*}

\begin{table*}[ht]
\centering
\caption{Classification accuracies (mean $\pm$ standard deviation) on six $128 \times 128$ ImageNet subsets for different utility losses, including gradient matching (GM)~\cite{zhao2020dataset} and distribution matching (DM)~\cite{Zhao_2023_WACV}, evaluated under a bpc budget of  $\le 192$ kB.}
\label{tab:my_table_4}
\renewcommand\arraystretch{1}
\setlength{\tabcolsep}{13pt}{
\begin{tabular}{lccccccc}
\toprule
Method & Nette & Woof & Fruit & Yellow & Meow & Squawk & Avg \\
\midrule 
\rowcolor{gray!30}
GM & & & & & & &\\
GM (Vanilla)~\cite{zhao2020dataset}) & 34.2{\scriptsize$\pm$1.7} & 22.5{\scriptsize$\pm$1.0} & 21.0{\scriptsize$\pm$0.9} & 37.1\textsuperscript{*}{\scriptsize$\pm$1.1}& 22.0{\scriptsize$\pm$0.6} & 32.0{\scriptsize$\pm$1.5} & 28.1 \\
GLaD~\cite{Cazenavette2023GeneralizingDD} & 35.4{\scriptsize$\pm$1.2} & 22.3{\scriptsize$\pm$1.1} & 20.7{\scriptsize$\pm$1.1} & --& 22.6{\scriptsize$\pm$0.8} & 33.8{\scriptsize$\pm$0.9} & 26.9 \\
H-GLaD~\cite{zhong2024hierarchical} & 36.9{\scriptsize$\pm$0.8} & 24.0{\scriptsize$\pm$0.8} & 22.4{\scriptsize$\pm$1.1} & -- & 24.1{\scriptsize$\pm$0.9} & 35.3{\scriptsize$\pm$1.0} & 28.5 \\
IDC~\cite{pmlr-v162-kim22c} & 45.4{\scriptsize$\pm$0.7} & 25.5{\scriptsize$\pm$0.7} & 26.8{\scriptsize$\pm$0.4} &--& 25.3{\scriptsize$\pm$0.6} & 34.6{\scriptsize$\pm$0.5} & 31.5 \\
FreD~\cite{shin2023frequency} & 49.1{\scriptsize$\pm$0.8} & 26.1{\scriptsize$\pm$1.1} & 30.0{\scriptsize$\pm$0.7} &--& 28.7{\scriptsize$\pm$1.0} & 39.7{\scriptsize$\pm$0.7} & 34.7 \\
DDiF~\cite{shin2025distilling} & \underline{61.2}{\scriptsize$\pm$1.0} & \textbf{35.2}{\scriptsize$\pm$1.7} & \textbf{37.8}{\scriptsize$\pm$1.1} & --&\underline{39.1}{\scriptsize$\pm$1.3} & \underline{54.3}{\scriptsize$\pm$1.0} & \underline{45.5} \\ 
\hline
GM-RUO (Ours) & \textbf{67.2}{\scriptsize$\pm$1.1} & \underline{33.1}{\scriptsize$\pm$1.1} & \underline{37.0}{\scriptsize$\pm$1.0} & \textbf{58.9}{\scriptsize$\pm$1.4}&\textbf{42.0}{\scriptsize$\pm$1.7} & \textbf{58.6}{\scriptsize$\pm$1.6} & \textbf{49.5} \\
\midrule
\rowcolor{gray!30}
DM & & & & & & &\\
DM (Vanilla)~\cite{Zhao_2023_WACV} & 30.4{\scriptsize$\pm$2.7} & 20.7{\scriptsize$\pm$1.0} & 20.4{\scriptsize$\pm$1.9} & 36.0\textsuperscript{*}{\scriptsize$\pm$0.8}& 20.1{\scriptsize$\pm$1.2} & 26.6{\scriptsize$\pm$2.6} & 25.7 \\
GLaD~\cite{Cazenavette2023GeneralizingDD} & 32.2{\scriptsize$\pm$1.7} & 21.2{\scriptsize$\pm$1.5} & 21.8{\scriptsize$\pm$1.8} & -- &22.3{\scriptsize$\pm$1.6} & 27.6{\scriptsize$\pm$1.9} & 25.0 \\
H-GLaD~\cite{zhong2024hierarchical} & 34.8{\scriptsize$\pm$1.0} & 23.9{\scriptsize$\pm$1.9} & 24.4{\scriptsize$\pm$2.1} & --&24.2{\scriptsize$\pm$1.1} & 29.5{\scriptsize$\pm$1.5} & 27.4 \\
IDC~\cite{pmlr-v162-kim22c} & 48.3{\scriptsize$\pm$1.3} & 27.0{\scriptsize$\pm$1.0} & 29.9{\scriptsize$\pm$0.7} & --&30.5{\scriptsize$\pm$1.0} & 38.8{\scriptsize$\pm$1.4} & 34.9 \\
FreD~\cite{shin2023frequency} & 56.2{\scriptsize$\pm$1.0} & 31.0{\scriptsize$\pm$1.2} & 33.4{\scriptsize$\pm$0.5} &--& 33.3{\scriptsize$\pm$0.6} & 42.7{\scriptsize$\pm$0.8} & 39.3 \\
DDiF~\cite{shin2025distilling} & \underline{69.2}{\scriptsize$\pm$1.0} & \underline{42.0}{\scriptsize$\pm$0.4} & \underline{45.3}{\scriptsize$\pm$1.8} & --&\underline{45.8}{\scriptsize$\pm$1.1} & \underline{64.6}{\scriptsize$\pm$1.1} & \underline{53.4} \\ 
\hline
DM-RUO (Ours) & \textbf{71.9}{\scriptsize$\pm$0.8} & \textbf{46.4}{\scriptsize$\pm$1.7} & \textbf{49.0}{\scriptsize$\pm$0.5} &\textbf{69.2}{\scriptsize$\pm$1.0} &\textbf{48.8}{\scriptsize$\pm$1.4} & \textbf{69.0}{\scriptsize$\pm$1.1} & \textbf{59.1} \\
\bottomrule
\end{tabular}}
\end{table*}

\begin{table}[t]
\centering
\caption{Wall-clock time of synthetic dataset generation.}
\label{tab:decoding_time}
\renewcommand\arraystretch{1}
\setlength{\tabcolsep}{15pt}{
\begin{tabular}{l c c c}
\toprule
Dataset&spc & Average Time (ms) & Time Per Sample (ms)\\
\midrule
\multirow{3}{*}{CIFAR-10}
    & 64   &  55.30 & 0.09 \\
    & 240  & 105.04 & 0.04 \\
    & 718  & 318.57 & 0.04 \\ \hline
\multirow{2}{*}{CIFAR-100}
    & 48   &  677.13 & 0.14 \\
    & 120  &  706.31 & 0.06 \\ \hline
\multirow{5}{*}{{\makecell{ImageNet Subset}}}  
    & 1    &  51.95 & 5.19 \\
    & 8    &  52.32 & 0.65 \\
    & 15   &  60.58 & 0.40 \\
    & 51   & 161.51 & 0.32 \\
    & 102  & 295.87 & 0.29 \\ 
\bottomrule
\end{tabular}
}
\end{table}

\subsection{Expanded Qualitative Results}
\begin{figure*}[ht]
\centering
\includegraphics[width=1.0\linewidth]{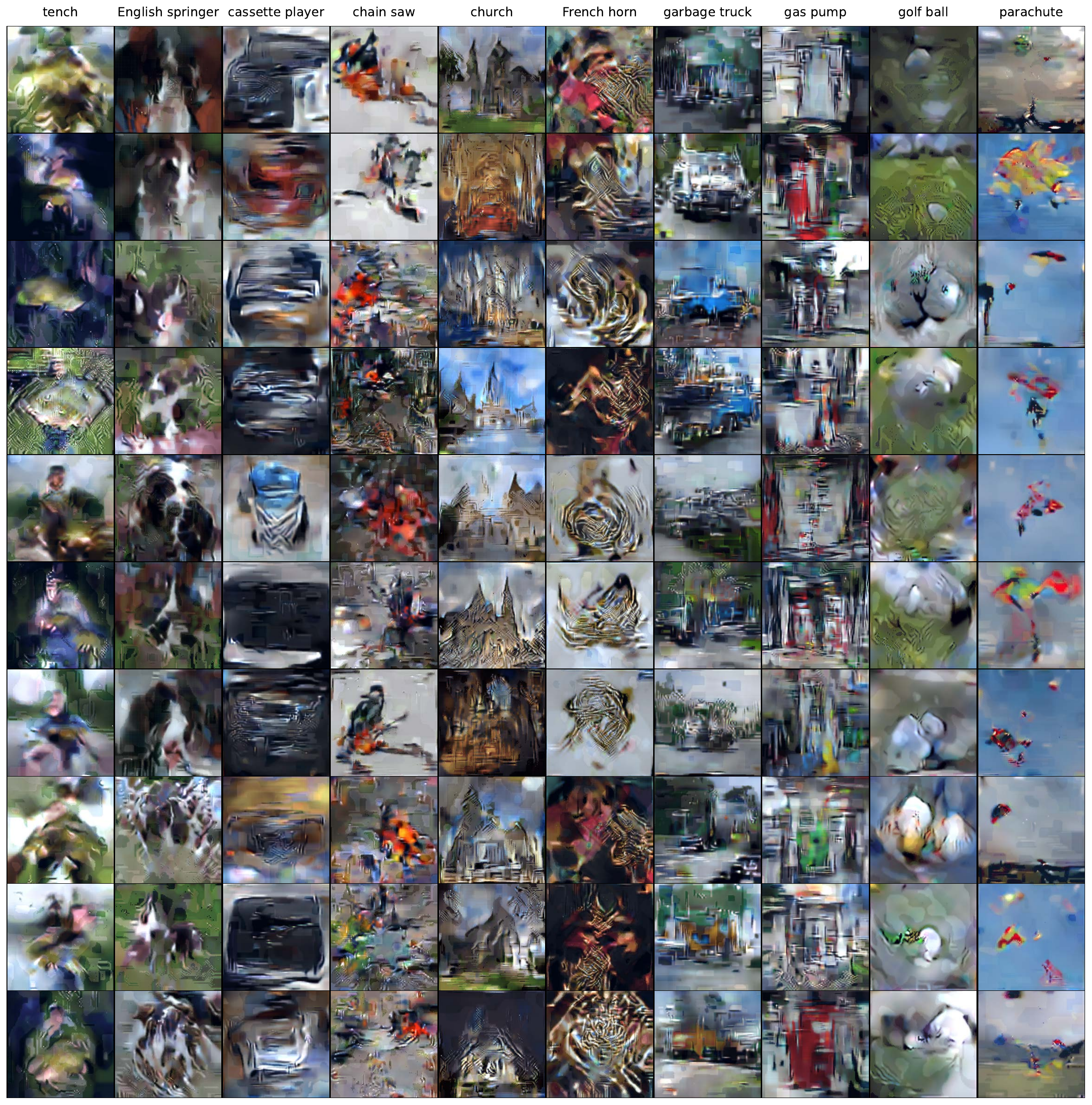}
\caption{Visualization of final synthetic samples on the Nette subset of ImageNet. }
\label{fig:full_nette}
\end{figure*}

\begin{figure*}[ht]
\centering
\includegraphics[width=1.0\linewidth]{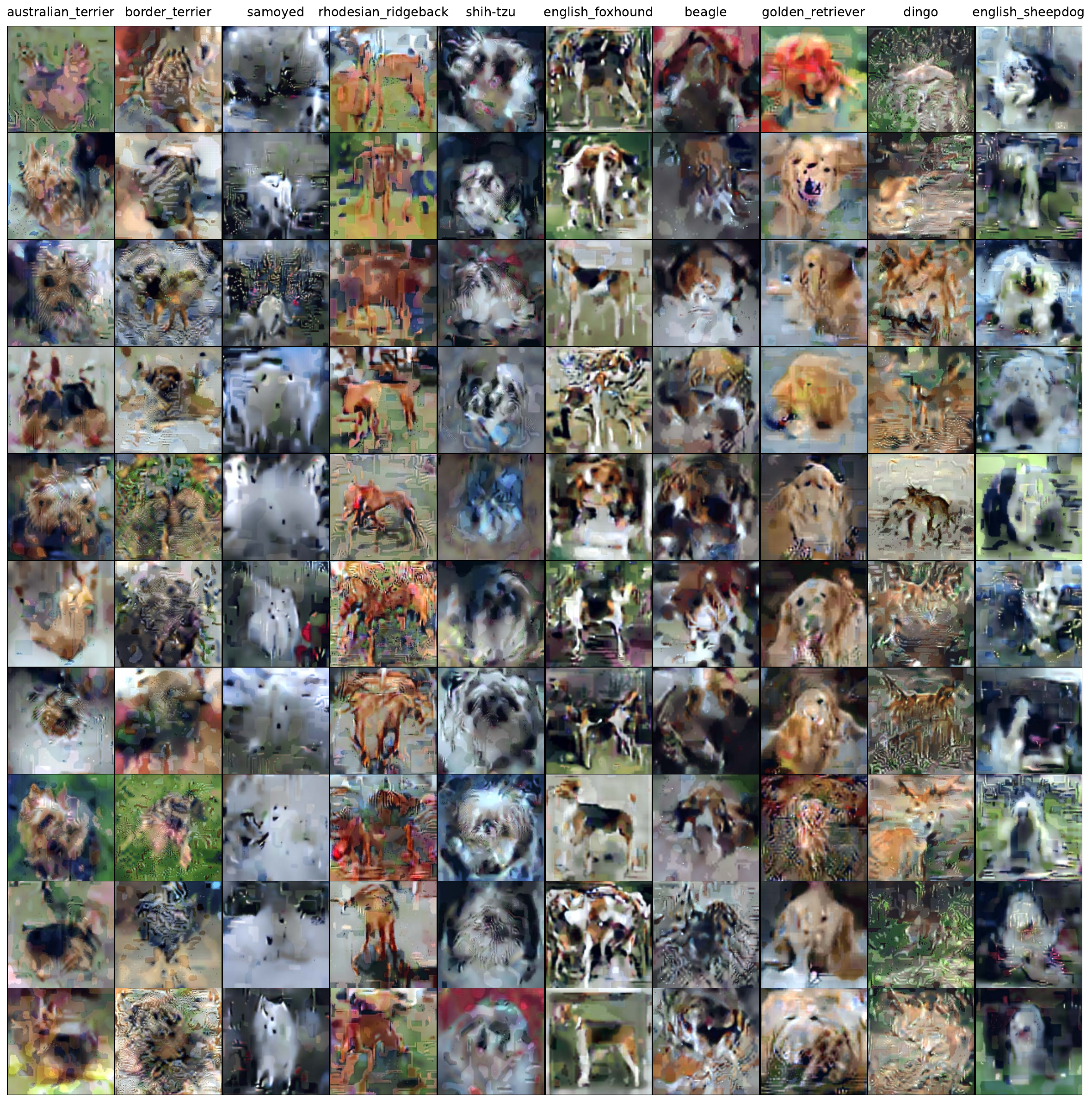}
\caption{Visualization of final synthetic samples on the Woof subset of ImageNet. }
\label{fig:full_woof}
\end{figure*}

\begin{figure*}[ht]
\centering
\includegraphics[width=1.0\linewidth]{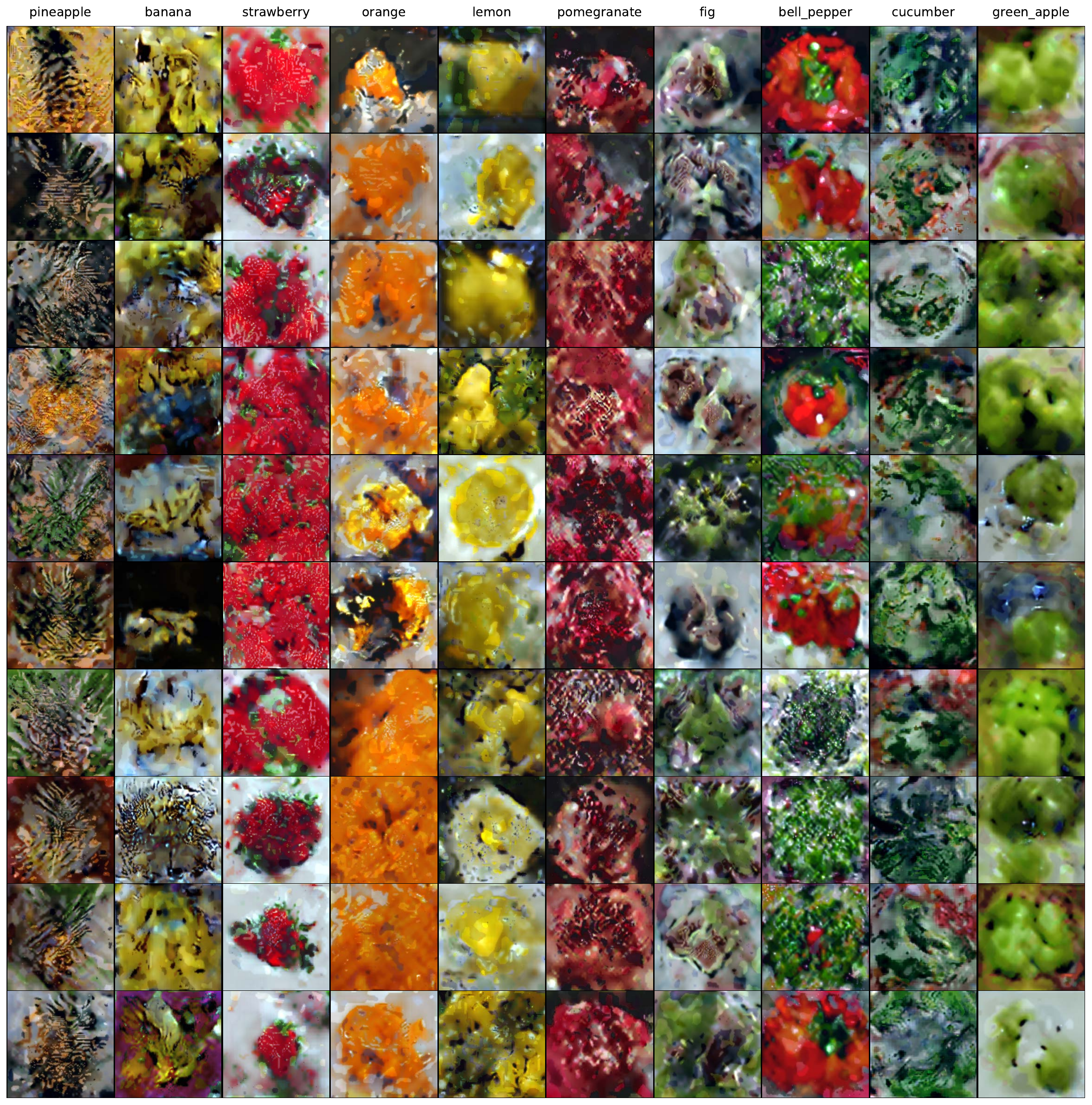}
\caption{Visualization of final synthetic samples on the Fruit subset of ImageNet. }
\label{fig:full_fruit}
\end{figure*}

\begin{figure*}[ht]
\centering
\includegraphics[width=1.0\linewidth]{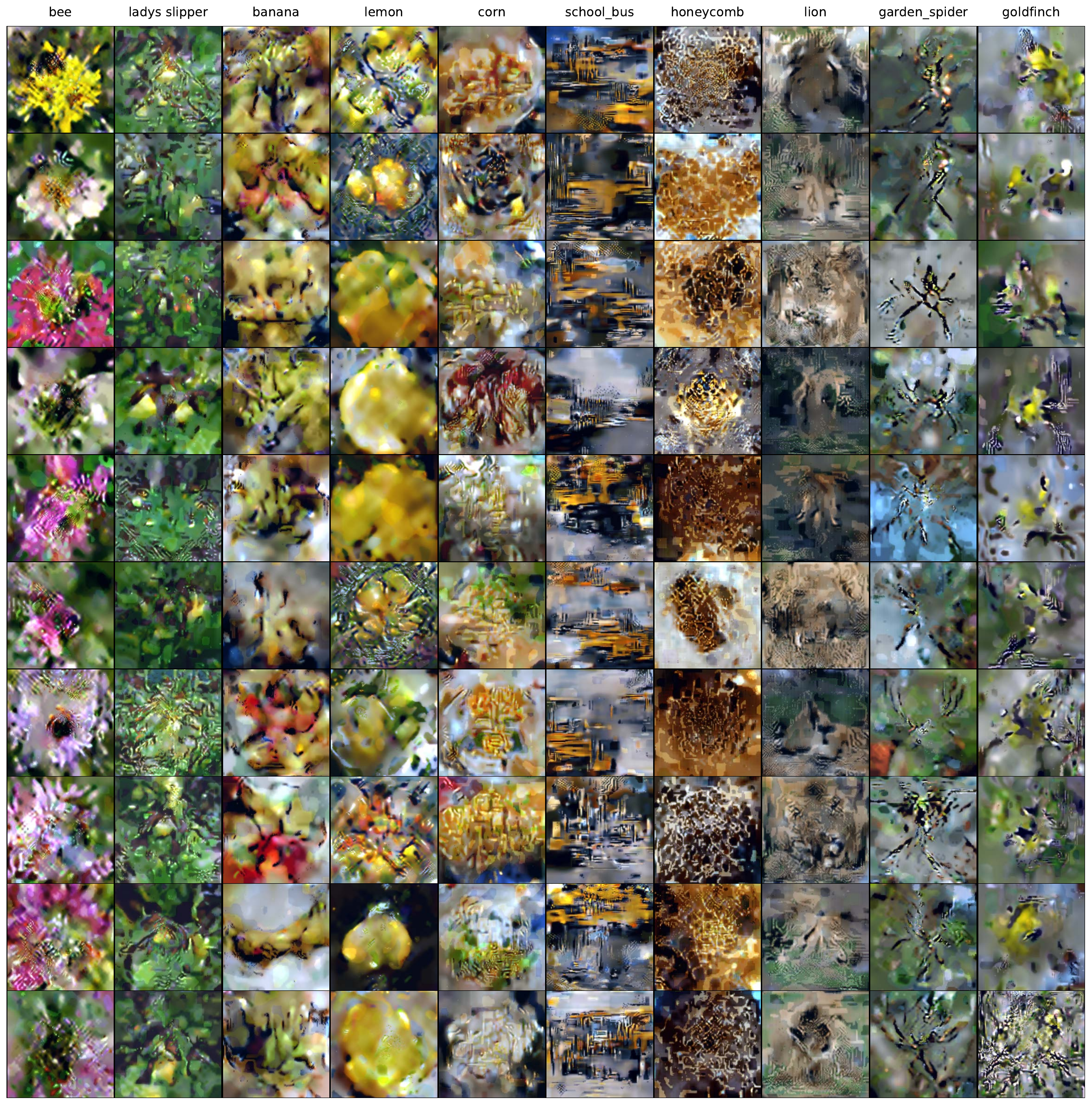}
\caption{Visualization of final synthetic samples on the Yellow subset of ImageNet. }
\label{fig:full_yellow}
\end{figure*}

\begin{figure*}[ht]
\centering
\includegraphics[width=1.0\linewidth]{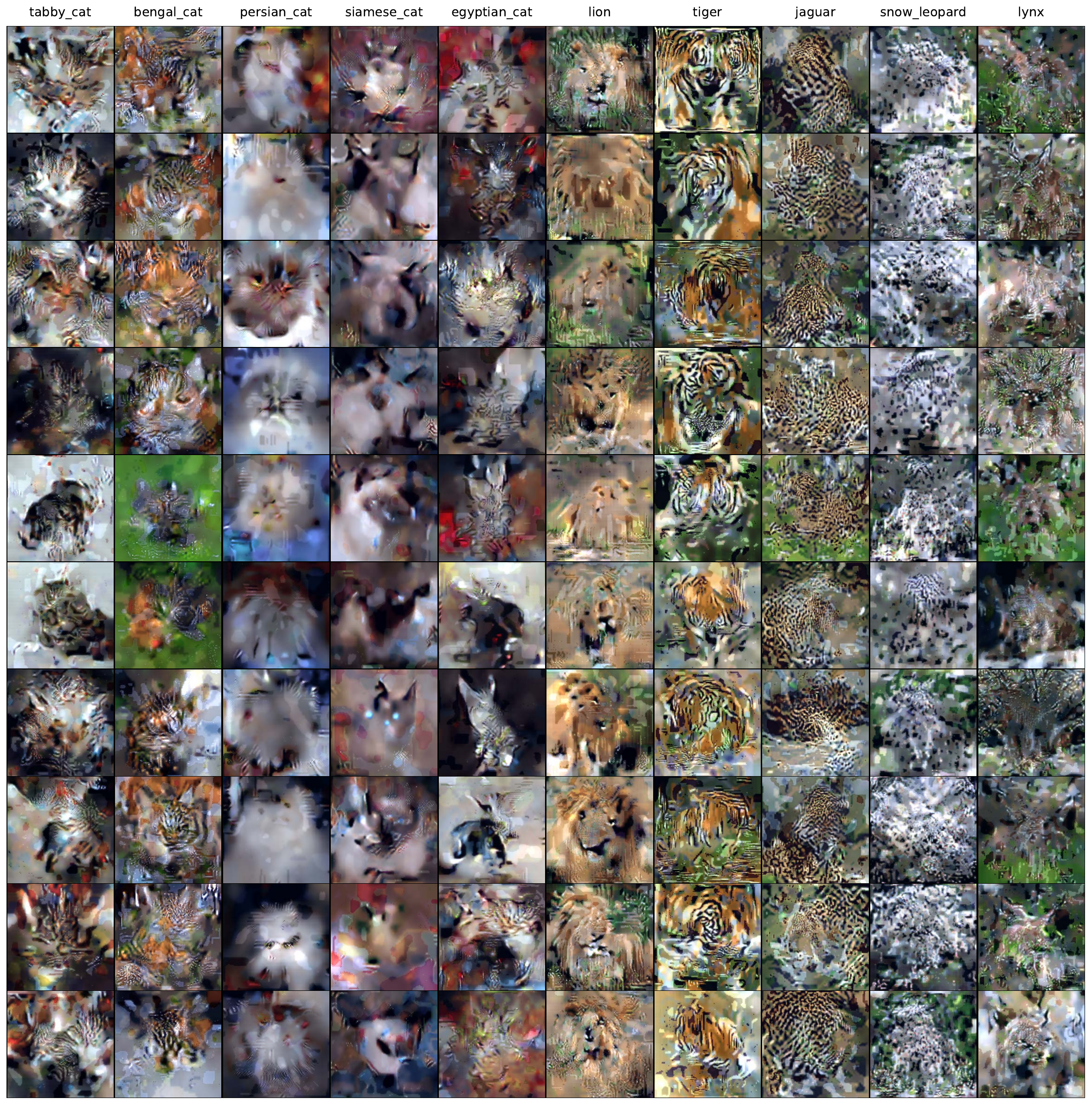}
\caption{Visualization of final synthetic samples on the Meow subset of ImageNet. }
\label{fig:full_meow}
\end{figure*}

\begin{figure*}[ht]
\centering
\includegraphics[width=1.0\linewidth]{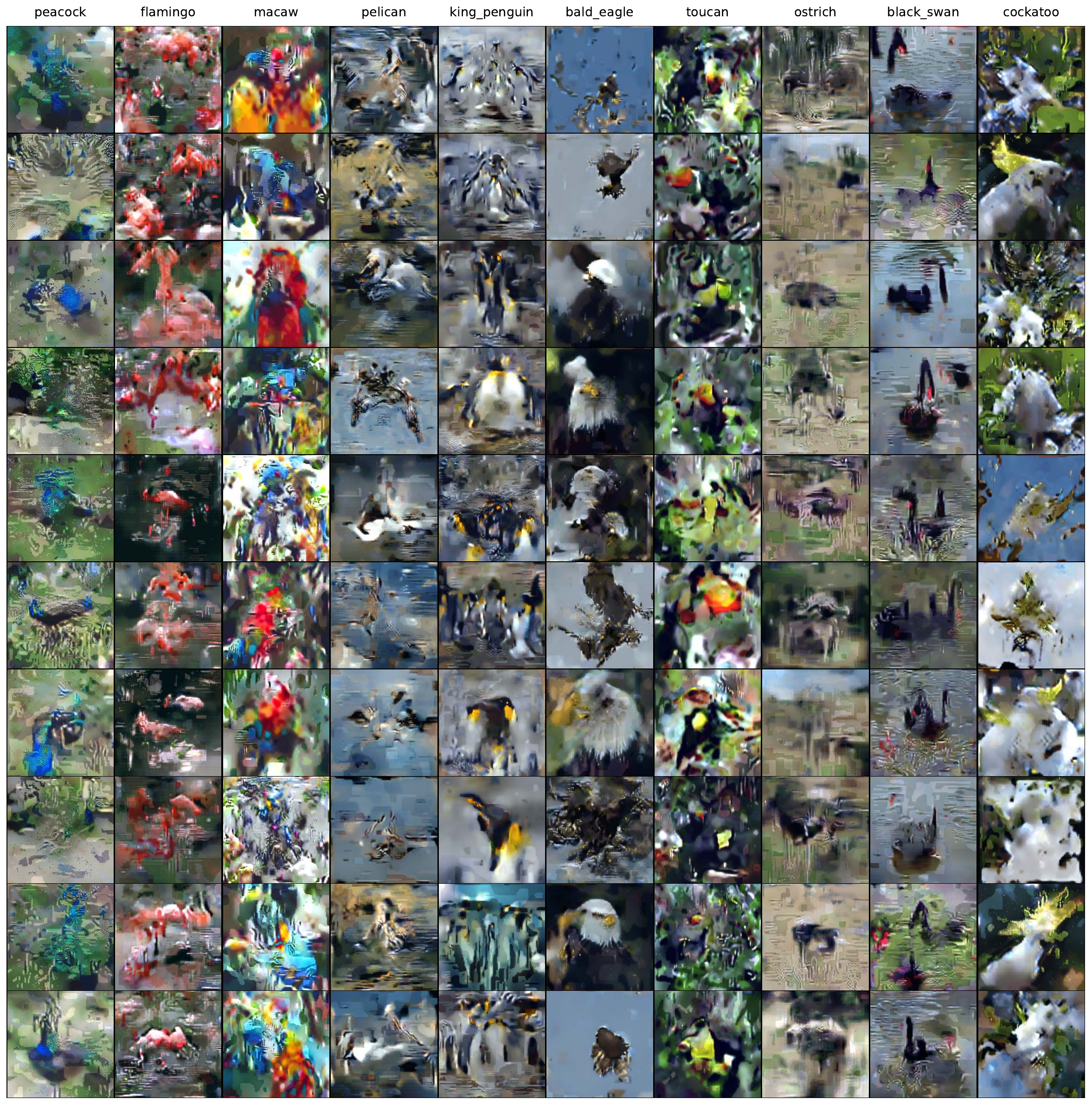}
\caption{Visualization of final synthetic samples on the Squawk subset of ImageNet. }
\label{fig:full_squawk}
\end{figure*}

\begin{figure*}[ht]
\centering
\includegraphics[width=1.0\linewidth]{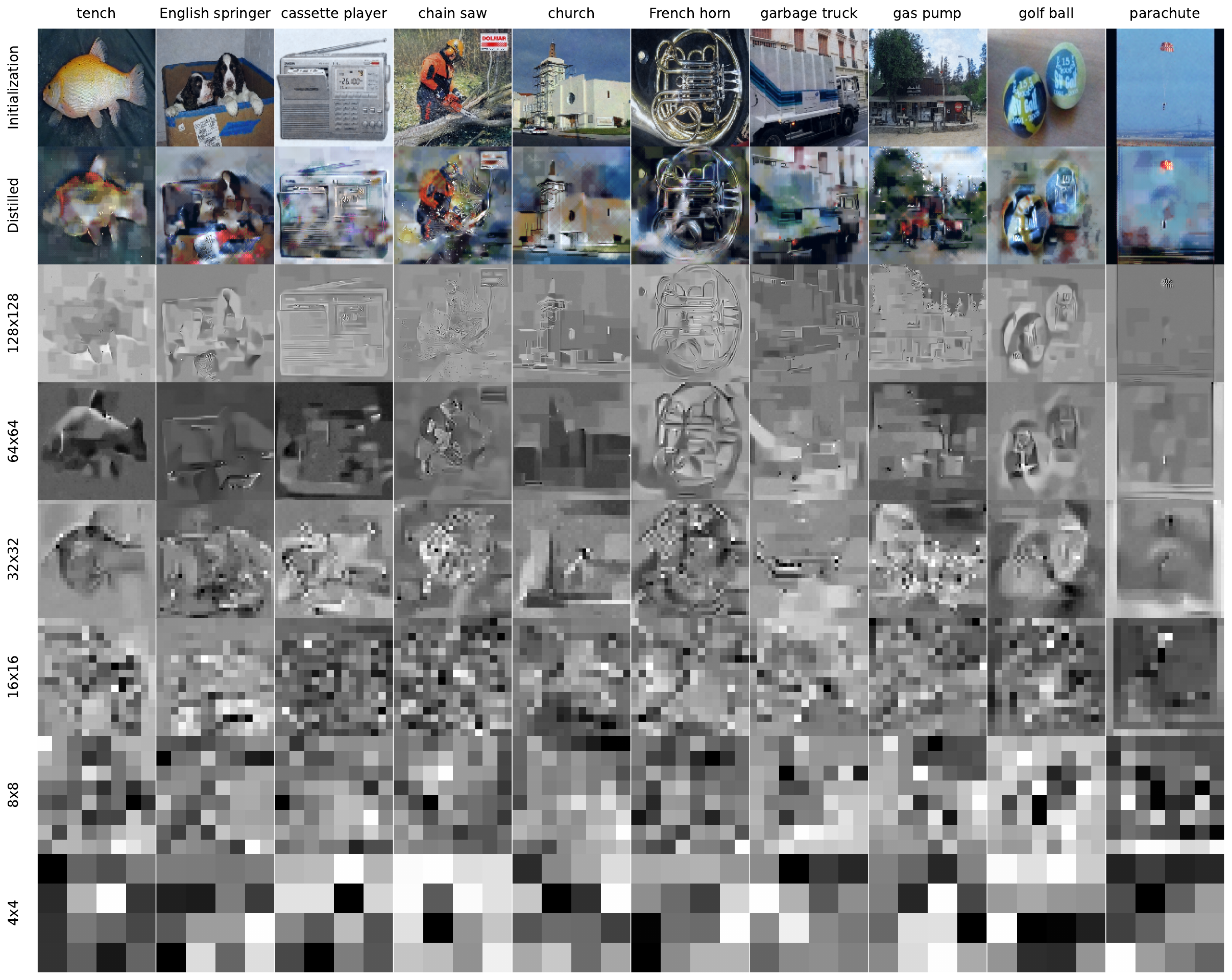}
\caption{Visualization of synthetic samples on the Nette subset of ImageNet using the GM loss.}
\label{fig:dc_nette}
\end{figure*}

\begin{figure*}[ht]
\centering
\includegraphics[width=1.0\linewidth]{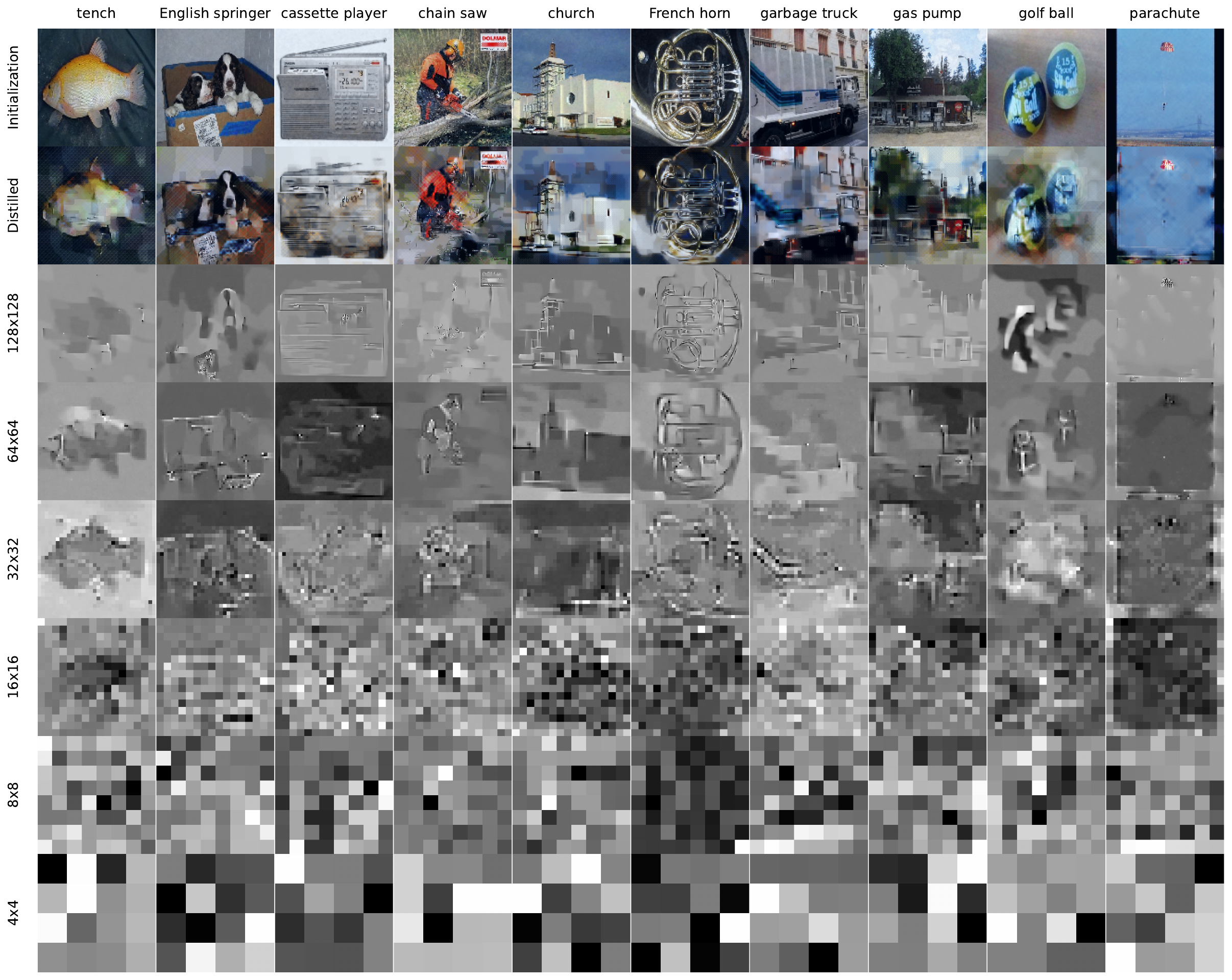}
\caption{Visualization of synthetic samples on the Nette subset of ImageNet using the DM loss.}
\label{fig:dm_nette}
\end{figure*}

\begin{figure*}[ht]
\centering
\includegraphics[width=1.0\linewidth]{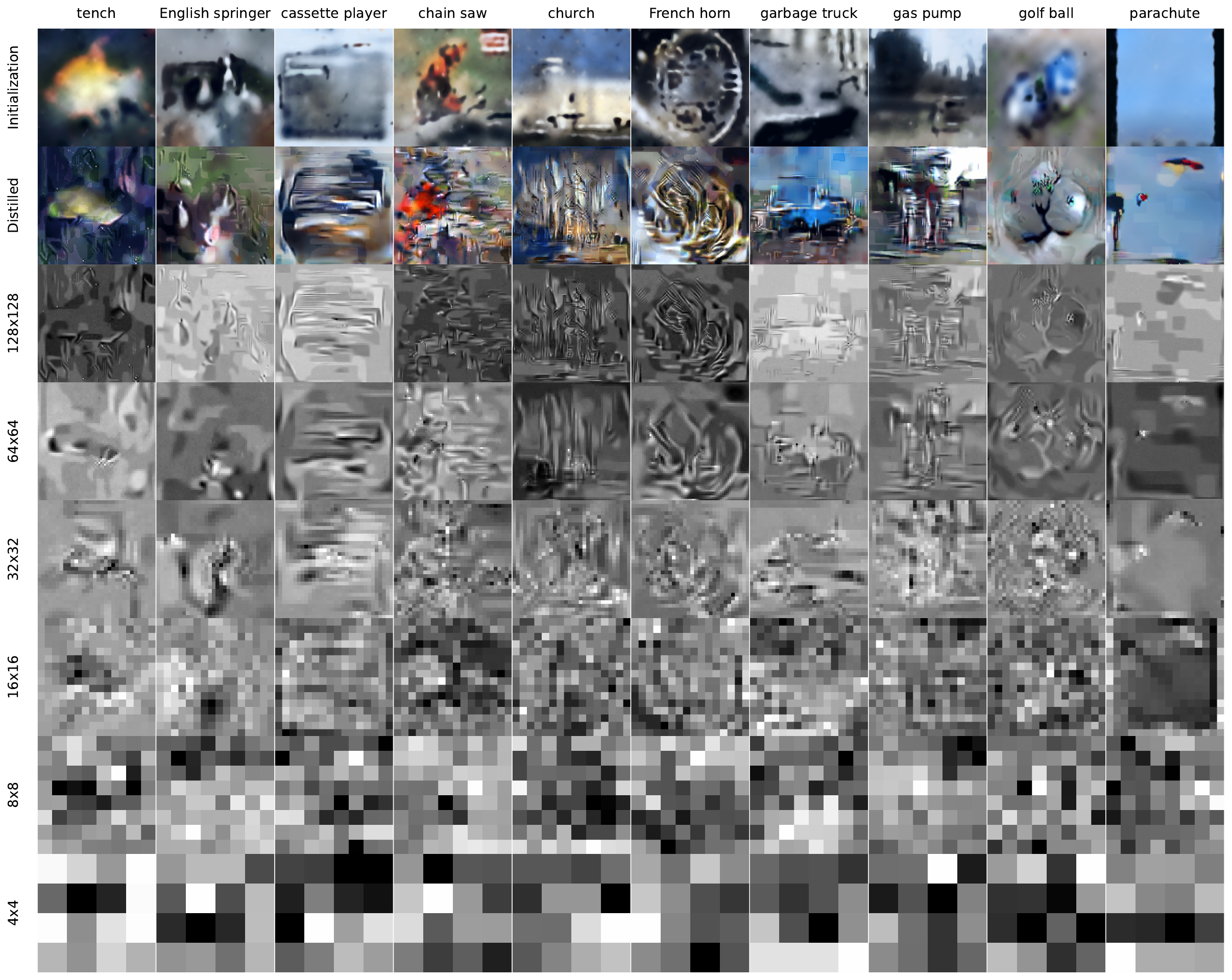}
\caption{Visualization of synthetic samples on the Nette subset of ImageNet using the TM loss.}
\label{fig:vis_nette}
\end{figure*}

\begin{figure*}[ht]
\centering
\includegraphics[width=1.0\linewidth]{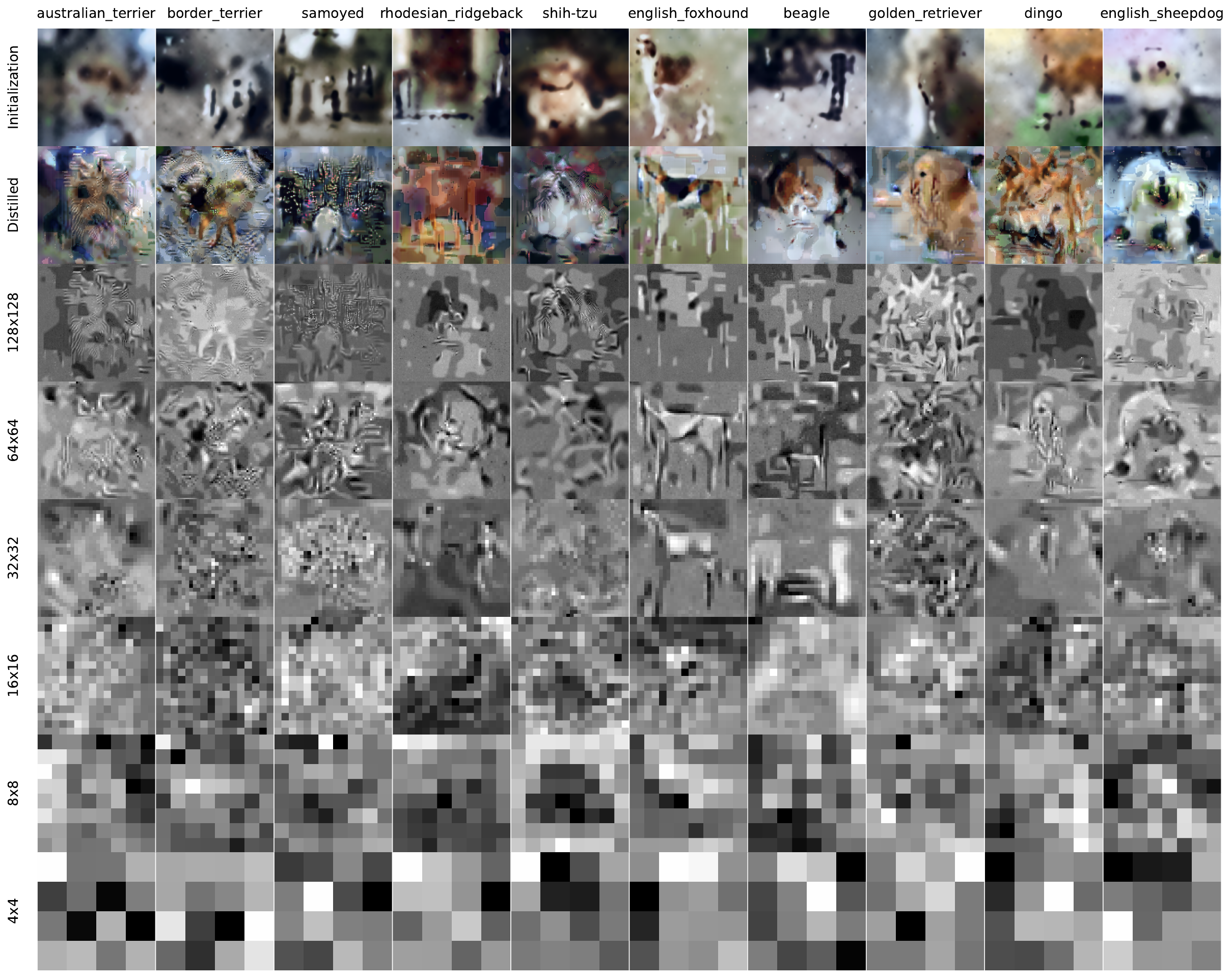}
\caption{Visualization of synthetic samples on the Woof subset of ImageNet using the TM loss.}
\label{fig:vis_woof}
\end{figure*}

\begin{figure*}[ht]
\centering
\includegraphics[width=1.0\linewidth]{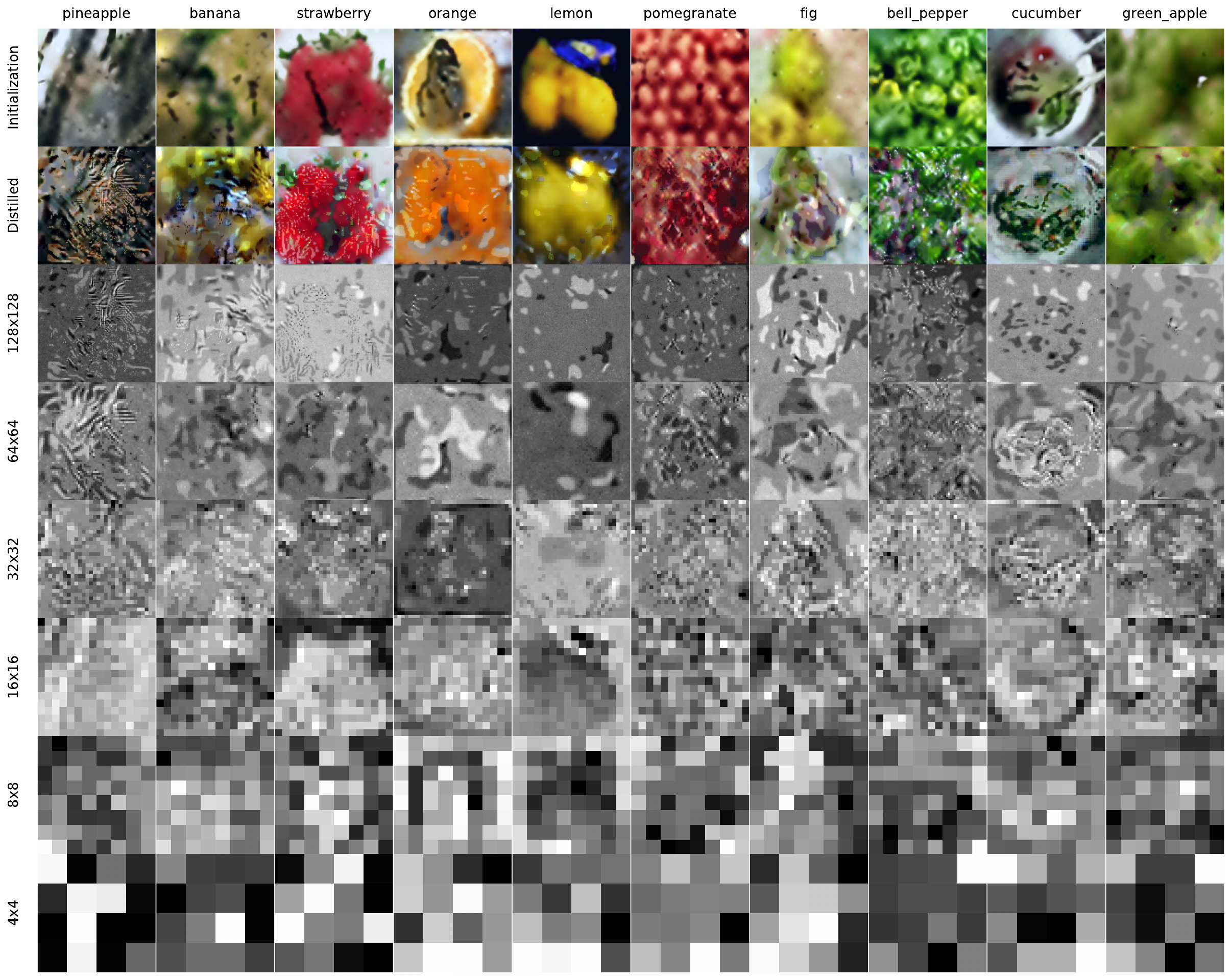}
\caption{Visualization of synthetic samples on the Fruit subset of ImageNet using the TM loss.}
\label{fig:vis_fruit}
\end{figure*}

\begin{figure*}[ht]
\centering
\includegraphics[width=1.0\linewidth]{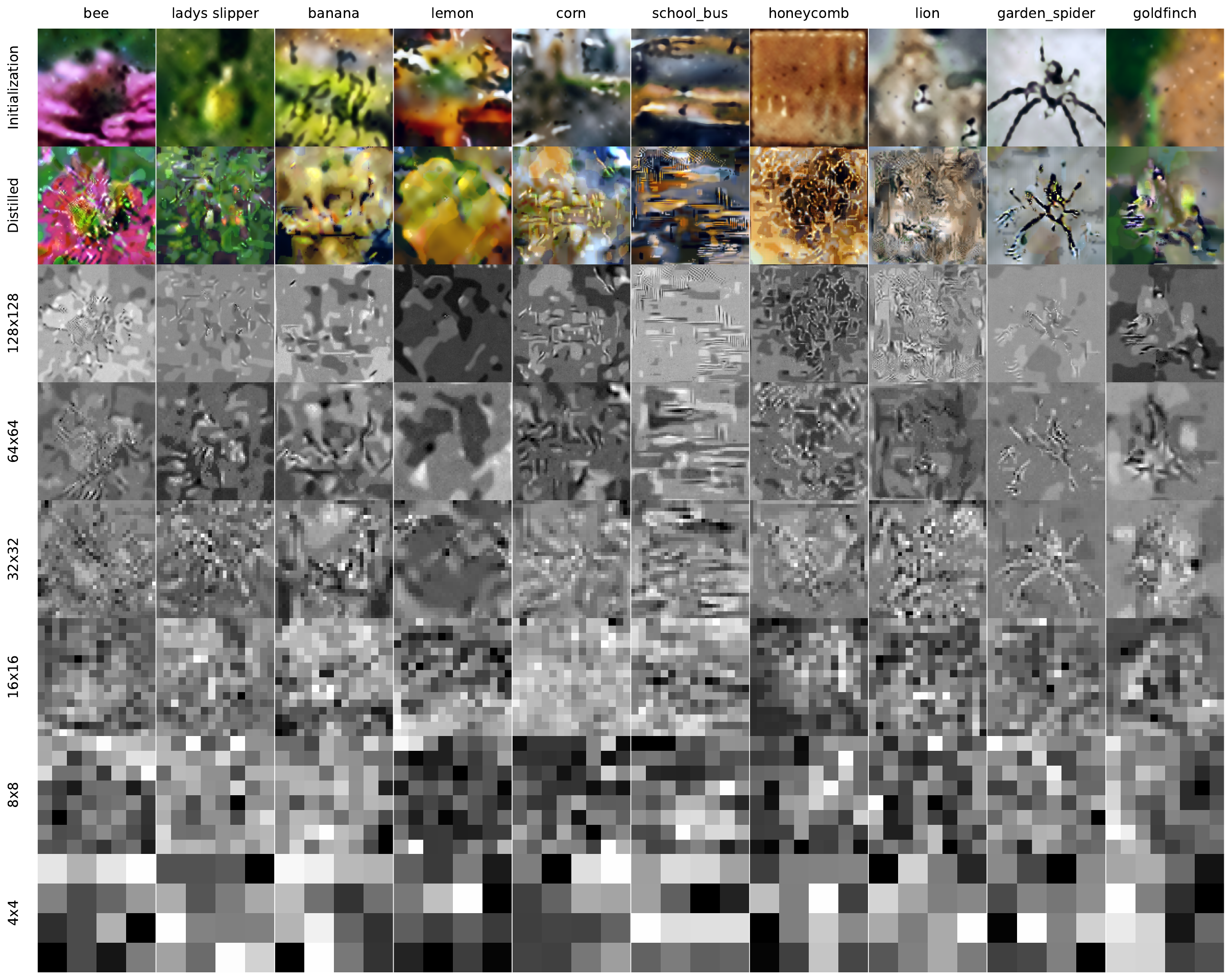}
\caption{Visualization of synthetic samples on the Yellow subset of ImageNet using the TM loss.}
\label{fig:vis_yellow}
\end{figure*}

\begin{figure*}[ht]
\centering
\includegraphics[width=1.0\linewidth]{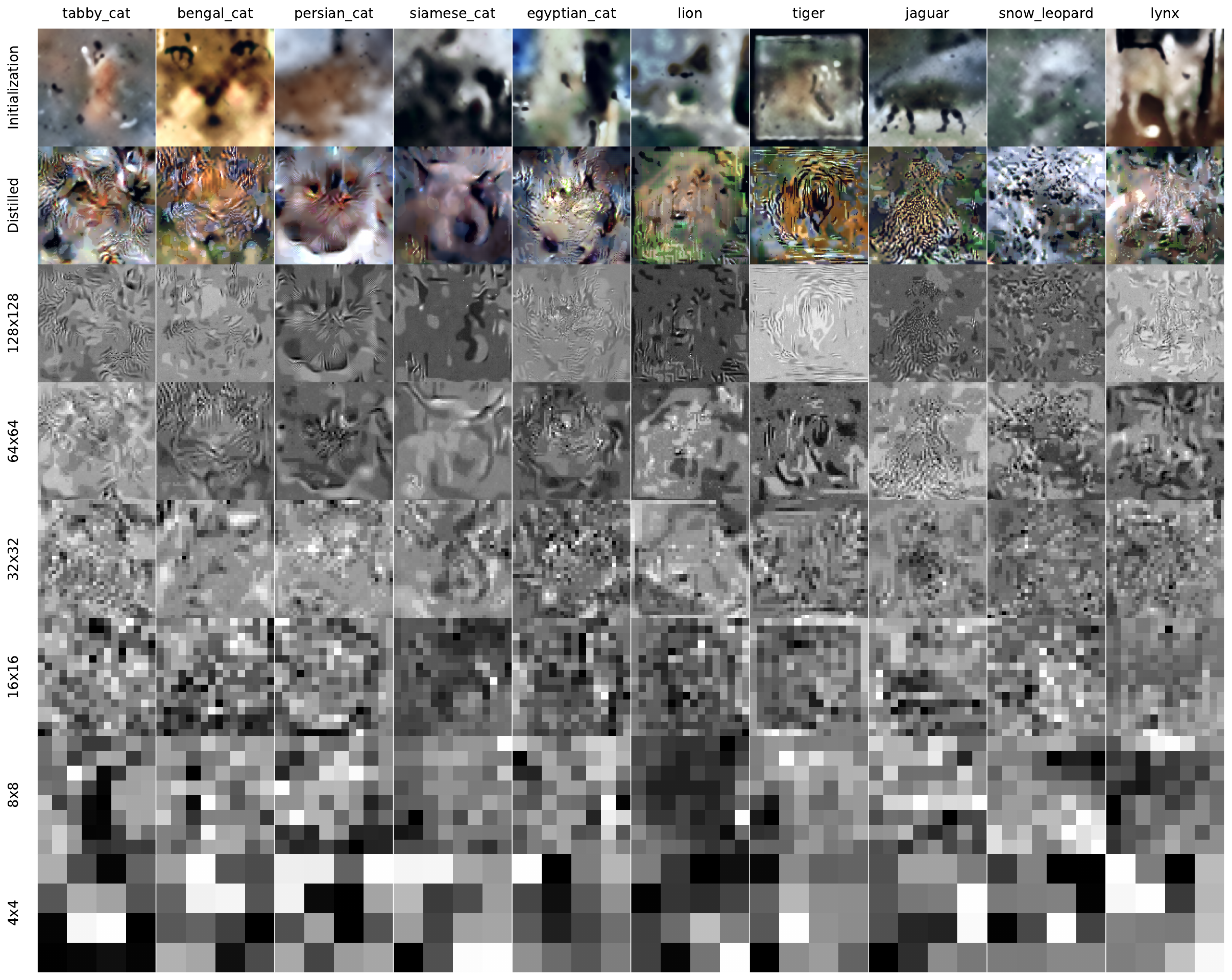}
\caption{Visualization of synthetic samples on the Meow subset of ImageNet using the TM loss.}
\label{fig:vis_meow}
\end{figure*}

\begin{figure*}[ht]
\centering
\includegraphics[width=1.0\linewidth]{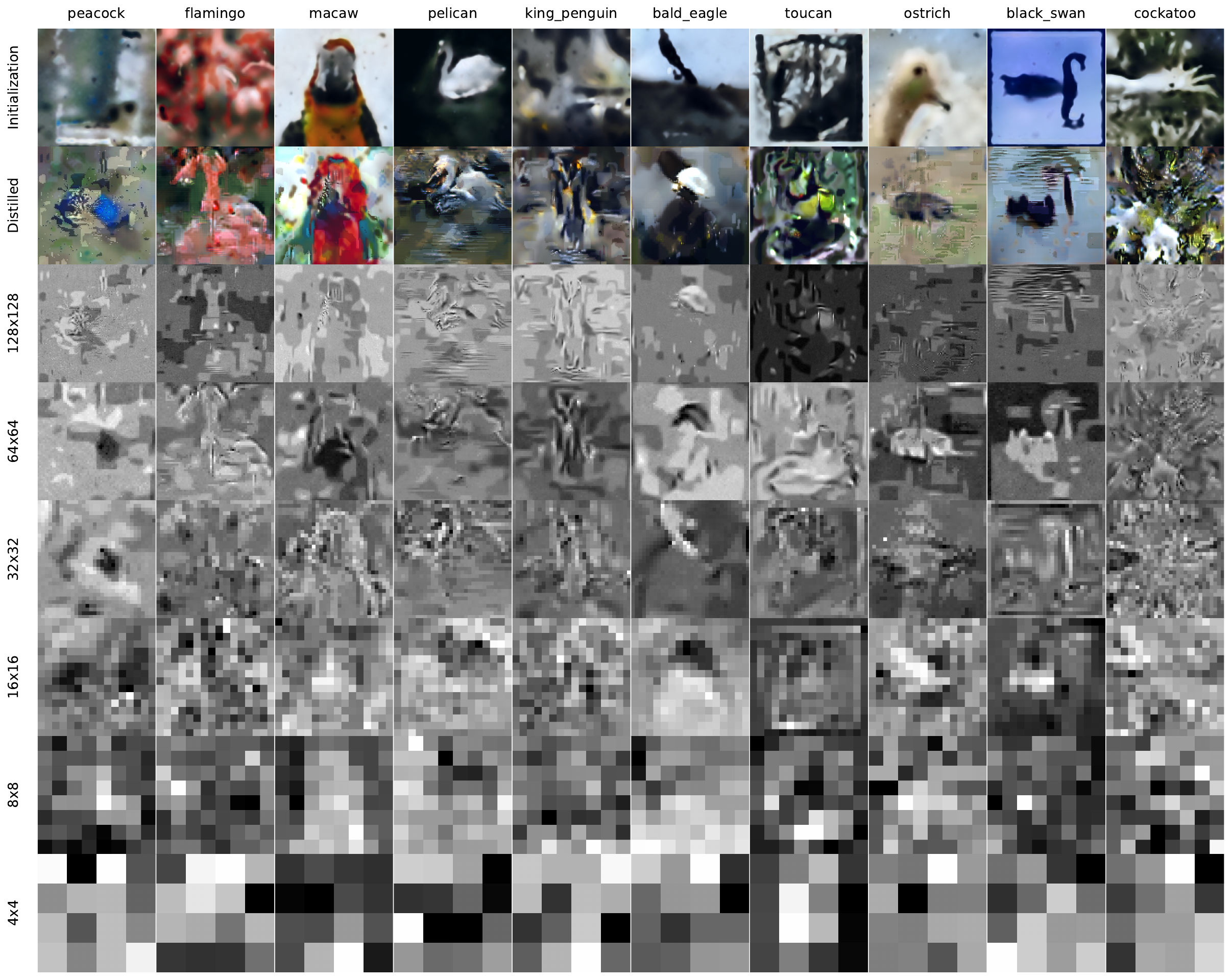}
\caption{Visualization of synthetic samples on the Squawk subset of ImageNet using the TM loss.}
\label{fig:vis_squawk}
\end{figure*}

Figs.~\ref{fig:full_nette} to~\ref{fig:full_squawk} showcase final synthetic samples optimized with the TM loss, whereas Figs.~\ref{fig:dc_nette} and~\ref{fig:dm_nette} juxtapose multiscale latent-code visualizations and their decoded reconstructions using the GM and DM objectives. Each panel’s first row presents initialization by ``overfitted'' image compression, the second row shows distilled images, and the last six rows display the corresponding multiscale latent codes. Finally, Figs.~\ref{fig:vis_nette} to~\ref{fig:vis_squawk} extend these visualizations across all six ImageNet subsets, demonstrating consistently structured latent representations and sample synthesis.

\end{document}